\documentclass[journal]{IEEEtran}
\usepackage[utf8]{inputenc} 
\usepackage[T1]{fontenc}    
\usepackage{hyperref}       
\usepackage{url}            
\usepackage{booktabs}       
\usepackage{nicefrac}       
\usepackage{microtype}      
\usepackage{amsthm}
\usepackage{amsbsy}
\usepackage{amsfonts,amssymb}
\usepackage{stmaryrd}
\usepackage{xcolor}
\usepackage{graphicx}
\usepackage{amsmath,mathtools}
\usepackage{algorithm,algorithmic}
\usepackage{subfigure}

\title{ORCCA: Optimal Randomized Canonical Correlation Analysis}

\author{Yinsong Wang and Shahin Shahrampour, {\it Senior Member}, {\it IEEE}
\thanks{Yinsong Wang and Shahin Shahrampour are with the Department of Mechanical and Industrial Engineering at Northeastern University, Boston, MA 02115 USA. Email addresses: {\tt\small \{wang.yinso,s.shahrampour\}@northeastern.edu}.}
}

\begin{document}
\newcommand\numberthis{\addtocounter{equation}{1}\tag{\theequation}}


\newcommand{\alg}{\text{ELSS}}


\newcommand{\0}{\mathbb{0}}
\newcommand{\1}{\mathbb{1}}

\newcommand{\E}{\mathbb{E}}
\newcommand{\R}{\mathbb{R}}
\renewcommand{\P}{\mathbb{P}}
\newcommand{\U}{\mathbb{U}}


\newcommand{\ab}{\mathbf{a}}
\newcommand{\bb}{\mathbf{b}}
\newcommand{\eb}{\mathbf{e}}
\newcommand{\ib}{\mathbf{i}}
\newcommand{\pb}{\mathbf{p}}
\newcommand{\qb}{\mathbf{q}}
\newcommand{\vb}{\mathbf{v}}
\newcommand{\ub}{\mathbf{u}}
\newcommand{\xb}{\mathbf{x}}
\newcommand{\yb}{\mathbf{y}}
\newcommand{\zb}{\mathbf{z}}

\newcommand{\Ab}{\mathbf{A}}
\newcommand{\Bb}{\mathbf{B}}
\newcommand{\Cb}{\mathbf{C}}
\newcommand{\Db}{\mathbf{D}}
\newcommand{\Eb}{\mathbf{E}}
\newcommand{\Fb}{\mathbf{F}}
\newcommand{\Gb}{\mathbf{G}}
\newcommand{\Hb}{\mathbf{H}}
\newcommand{\Ib}{\mathbf{I}}
\newcommand{\Jb}{\mathbf{J}}
\newcommand{\Kb}{\mathbf{K}}
\newcommand{\Lb}{\mathbf{L}}
\newcommand{\Pb}{\mathbf{P}}
\newcommand{\Qb}{\mathbf{Q}}
\newcommand{\Rb}{\mathbf{R}}
\newcommand{\Sb}{\mathbf{S}}
\newcommand{\Vb}{\mathbf{V}}
\newcommand{\Ub}{\mathbf{U}}
\newcommand{\Wb}{\mathbf{W}}
\newcommand{\Xb}{\mathbf{X}}
\newcommand{\Yb}{\mathbf{Y}}
\newcommand{\Zb}{\mathbf{Z}}

\newcommand{\alphab}{\boldsymbol{\alpha}}
\newcommand{\betab}{\boldsymbol{\beta}}
\newcommand{\gammab}{\boldsymbol{\gamma}}
\newcommand{\phib}{\boldsymbol{\phi}}
\newcommand{\Phib}{\boldsymbol{\Phi}}
\newcommand{\Qhib}{\boldsymbol{\Qhi}}
\newcommand{\omegab}{\boldsymbol{\omega}}
\newcommand{\psib}{\boldsymbol{\psi}}
\newcommand{\sigmab}{\boldsymbol{\sigma}}
\newcommand{\nub}{\boldsymbol{\nu}}
\newcommand{\thetab}{\boldsymbol{\theta}}
\newcommand{\delb}{\boldsymbol{\delta}}
\newcommand{\rhob}{\boldsymbol{\rho}}
\newcommand{\Pib}{\boldsymbol{\Pi}}
\newcommand{\pib}{\boldsymbol{\pi}}
\newcommand{\Sigmab}{\boldsymbol{\Sigma}}


\newcommand{\Cc}{\mathcal{C}}
\newcommand{\Ec}{\mathcal{E}}
\newcommand{\Fc}{\mathcal{F}}
\newcommand{\Hc}{\mathcal{H}}
\newcommand{\Lc}{\mathcal{L}}
\newcommand{\Nc}{\mathcal{N}}
\newcommand{\Oc}{\mathcal{O}}
\newcommand{\Pc}{\mathcal{P}}
\newcommand{\Rc}{\mathcal{R}}
\newcommand{\Uc}{\mathcal{U}}
\newcommand{\Xc}{\mathcal{X}}
\newcommand{\Yc}{\mathcal{Y}}


\newcommand{\tauh}{\widehat{\tau}}
\newcommand{\Sigmah}{\widehat{\Sigma}}

\newcommand{\fh}{\widehat{f}}
\newcommand{\gh}{\widehat{g}}
\newcommand{\kh}{\widehat{k}}
\newcommand{\qh}{\widehat{q}}
\newcommand{\Rh}{\widehat{R}}


\newcommand{\alphabh}{\widehat{\boldsymbol{\alpha}}}
\newcommand{\thetabh}{\widehat{\boldsymbol{\theta}}}

\newcommand{\qbh}{\widehat{\mathbf{q}}}

\newcommand{\Kbh}{\widehat{\mathbf{K}}}


\newcommand{\Fch}{\widehat{\mathcal{F}}}


\newcommand{\argmin}{\text{argmin}}
\newcommand{\arginf}{\text{arginf}}
\newcommand{\argmax}{\text{argmax}}
\newcommand{\minimize}{\text{minimize}}
\newcommand{\maximize}{\text{maximize}}
\newcommand{\supp}{\text{supp}}


\newcommand{\TV}{\text{TV}}
\newcommand{\norm}[1]{\left\lVert#1\right\rVert}
\newcommand{\tr}[1]{\text{Tr}\left[#1\right]}
\newcommand{\inn}[1]{\left<#1\right>}
\newcommand{\seal}[1]{\left \lceil #1\right \rceil}
\newcommand{\floor}[1]{\left \lfloor #1\right \rfloor}
\newcommand{\abs}[1]{\left|#1\right|}
\newcommand{\ind}[1]{\mathbf{1}\left(#1\right)}
\newcommand{\ex}[1]{\E\left[#1\right]}


\newtheorem{theorem}{Theorem}
\newtheorem{acknowledgement}[theorem]{Acknowledgement}
\newtheorem{assumption}{Assumption}
\newtheorem{conjecture}[theorem]{Conjecture}
\newtheorem{corollary}[theorem]{Corollary}
\newtheorem{definition}{Definition}
\newtheorem{example}{Example}
\newtheorem{lemma}[theorem]{Lemma}
\newtheorem{fact}{Fact}
\newtheorem{problem}{Problem}
\newtheorem{proposition}[theorem]{Proposition}
\newtheorem{remark}{Remark}
\newtheorem{solution}[theorem]{Solution}
\newtheorem{summary}[theorem]{Summary}

\newcommand{\bl}{\color{blue}}

\maketitle

\begin{abstract}
Random features approach has been widely used for kernel approximation in large-scale machine learning. A number of recent studies have explored data-dependent sampling of features, modifying the stochastic oracle from which random features are sampled. While proposed techniques in this realm improve the approximation, their suitability is often verified on a single learning task. In this paper, we propose a {\it task-specific} scoring rule for selecting random features, which can be employed for different applications with some adjustments. We restrict our attention to Canonical Correlation Analysis (CCA), and we provide a novel, principled guide for finding the score function maximizing the canonical correlations. We prove that this method, called ORCCA, can outperform (in expectation) the corresponding Kernel CCA with a default kernel. Numerical experiments verify that ORCCA is significantly superior than other approximation techniques in the CCA task.
\end{abstract}


\section{Introduction}

Kernel methods are powerful tools to capture the nonlinear representation of data by mapping the dataset to a high-dimensional feature space. Despite their tremendous success in various machine learning problems, kernel methods suffer from massive computational cost on large datasets. The time cost of computing the kernel matrix alone scales quadratically with data, and if the learning method involves inverting the matrix (e.g., kernel ridge regression), the cost would increase to cubic. This computational bottleneck motivated a great deal of research on kernel approximation, where the seminal work of \cite{rahimi2008random} on {\it random features} is a prominent point in case. For the class of shift-invariant kernels, they showed that one can approximate the kernel by Monte-Carlo sampling from the inverse Fourier transform of the kernel. This idea has been used in solving many machine learning problems, including distributed learning \cite{richards2020decentralised, wang2020distributed}, online learning \cite{hu2015dependent,nguyen2017large}, and deep learning \cite{huang2013random}, etc.

Due to the practical success of random features, the idea was later used for one of the ubiquitous problems in statistics and machine learning, namely Canonical Correlation Analysis (CCA). CCA derives a pair of linear mappings of two datasets, such that the correlation between the projected datasets is maximized. Similar to other machine learning methods, CCA also has a nonlinear counterpart called Kernel Canonical Correlation Analysis (KCCA) \cite{lai2000kernel}, which provides a more flexible framework for maximizing the correlation. Due to the prohibitive computational cost of KCCA, Randomized Canonical Correlation Analysis (RCCA) was introduced \cite{lopez2013randomized,lopez2014randomized} to serve as a surrogate for KCCA. RCCA uses random features for transformation of the two datasets. Therefore, it provides the flexibility of nonlinear mappings with a moderate computational cost.

On the other hand, more recently, {\it data-dependent} sampling of random features has been an intense focus of research in the machine learning community. The main objective is to modify the stochastic oracle from which random features are sampled to improve a certain performance metric. Examples include \cite{yang2015carte,chang2017data} with a focus only on kernel approximation as well as \cite{sinha2016learning,avron2017random,bullins2017not} with the goal of better generalization in supervised learning. While the proposed techniques in this realm improve their respective learning tasks, they are not necessarily suitable for other learning tasks, such as CCA which is the focus of this work.  

In this paper, we propose a task-specific scoring rule for re-weighting random features, which can be employed for various applications with some adjustments. In particular, our scoring rule depends on a matrix that can be adjusted based on the application. We first observe that a number of data-dependent sampling methods (e.g., leverage scores in \cite{avron2017random} and energy-based sampling in \cite{shahrampour2018data}) can be recovered by our scoring rule using specific choices of the matrix. Then, we draw a connection between the scoring rule and correlation analysis/dimension reduction problems that deal with trace optimization objectives. As an important case study, we focus on CCA and provide a principled guide for finding the score function maximizing the canonical correlations. Our result reveals a novel data-dependent method for selecting features, called Optimal Randomized Canonical Correlation Analysis (ORCCA). This suggests that prior data-dependent methods are not necessarily optimal for the CCA task. We also prove that ORCCA achieves a better performance compared to KCCA (in expectation) with a default kernel. We conduct extensive numerical experiments verifying that ORCCA indeed introduces significant improvement over the state-of-the-art in random features for CCA.

The rest of this paper is organized as follows. In Section \ref{sec2}, we provide the preliminaries on random features, canonical correlation analysis and formally define the problem we will address in this paper. This section also includes the related literature. In Section \ref{sec:generalscore}, we propose our score function, discuss its connection with existing score functions in supervised learning, and show a class of problems that is compatible with our score function. In Section \ref{sec:CCA}, we present our theoretical results. We illustrate the effectiveness of ORCCA on benchmark datasets in Section \ref{simulation} and conclude in Section \ref{Conclusion}.

\section{Preliminaries and Problem Setting}\label{sec2}

{\bf Notation:} We denote by $[n]$ the set of positive integers $\{1,\ldots,n\}$, by $\tr{\cdot}$ the trace operator, by $\inn{\cdot,\cdot}$ the standard inner product, by $\norm{\cdot}$ the spectral (respectively, Euclidean) norm of a matrix (respectively, vector), and by $\ex{\cdot}$ the expectation operator. Boldface lowercase variables (e.g., $\ab$) are used for vectors, and boldface uppercase variables (e.g., $\Ab$) are used for matrices. $[\Ab]_{ij}$ denotes the $ij$-th entry of matrix $\Ab$. The vectors are all in column form. 
 
\subsection{Random Features and Kernel Approximation}\label{sec:random}
Kernel methods are powerful tools for data representation, commonly used in various machine learning problems. Let $\{\xb_i\}_{i=1}^n$ be a set of given points where $\xb_i \in \Xc \subseteq \R^{d_x}$ for any $i\in[n]$, and consider a symmetric positive-definite function $k(\cdot,\cdot)$ such that $\sum^n_{i,j=1}\alpha_i \alpha_j k(\xb_i,\xb_j)\geq 0$ for $\alphab\in \R^n$. Then, $k(\cdot,\cdot)$ is called a positive (semi-)definite kernel, serving as a similarity measure between any pair of vectors $(\xb_i,\xb_j)$. This class of kernels can be thought as inner product of two vectors that map the points from a $d_x$-dimensional space to a higher dimensional space (and potentially infinite-dimensional space).

Despite the widespread use of kernel methods in machine learning, they have an evident computational issue. Computing the kernel for every pair of points costs $O(n^2)$, and if the learning method requires inverting that matrix (e.g., kernel ridge regression), the cost would increase to $O(n^3)$. This particular disadvantage makes kernel method impractical for large-scale machine learning. 

An elegant method to address this issue was the use of random Fourier features for kernel approximation \cite{rahimi2008random}. Let $p(\omegab)$ be a probability density with support $\Omega \subseteq \R^{d_x}$. Consider any kernel function in the following form with a corresponding feature map $\phi(\xb,\omegab)$, such that
\begin{equation}\label{kernel function}
    \begin{aligned}
    k(\xb,\xb') &= \int_\Omega \phi(\xb,\omegab)\phi(\xb',\omegab)p(\omegab)d\omegab\\
    &\approx \frac{1}{M} \sum_{m=1}^M \phi(\xb,\omegab_m)\phi(\xb',\omegab_m),
    \end{aligned}
\end{equation}
where $\{\omegab_m\}_{m=1}^M$ are independent samples from $p(\omegab)$, called {\it random features}. Examples of kernels taking the form \eqref{kernel function} include shift-invariant kernels \cite{rahimi2008random} or dot product (e.g., polynomial) kernels \cite{kar2012random} (see Table 1 in \cite{yang2014random} for an exhaustive list). Let us now define
\begin{align}\label{zvector}
\zb(\omegab)\triangleq[\phi(\xb_1,\omegab),\ldots,\phi(\xb_n,\omegab)]^\top.
\end{align}
Then, the kernel matrix $[\Kb]_{ij}=k(\xb_i,\xb_j)$ can be approximated with $\Zb\Zb^\top$ where $\Zb\in \R^{n\times M}$ is defined as
\begin{align}\label{trasformed matrix}
\Zb &\triangleq \frac{1}{\sqrt{M}}[\zb(\omegab_1),\ldots,\zb(\omegab_M)].
\end{align}
The low-rank approximation above can save significant computational cost when $M \ll n$. As an example, for kernel ridge regression the time cost would reduce from $O(n^3)$ to $O(nM^2)$. Since the main motivation of using randomized features is to reduce the computational cost of kernel methods (with $M \ll n$), this observation will naturally raise the following question:
\begin{problem}(Informal)\label{P1}
Can we develop a sampling (or selection) mechanism for random features that takes into account the ``learning task'' to improve the performance compared to plain sampling? 
\end{problem}
Section \ref{sec:generalscore} will shed light on Problem \ref{P1}. First, in Section \ref{sec:score}, we will propose a score function with a potential to be adapted to different learning tasks. Section \ref{sec:relation} will show the connection between the proposed score function and two existing score functions for random features in supervised learning. Then, Section \ref{sec:LinktoDR} will provide another class of problems (i.e., dimensionality reduction / correlation analysis) for which the proposed score function can prove useful. In this paper, we focus on kernel CCA (as one potential application), introduce our main question in Problem \ref{P2}, and provide our theoretical results.   

\subsection{Overview of Canonical Correlation Analysis}\label{sec:ccaoverview}
Linear CCA was introduced in \cite{hotelling1936relations}  as a method of correlating linear relationships between two multi-dimensional random variables $\Xb=[\xb_1,\ldots,\xb_n]^\top\in \R^{n\times d_x}$ and $\Yb=[\yb_1,\ldots,\yb_n]^\top\in \R^{n\times d_y}$. This problem is often formulated as finding a pair of canonical  bases $\Pib_x$ and $\Pib_y$ such that $\|\text{corr}(\Xb\Pib_x,\Yb\Pib_y)-\Ib_r\|_F$ is minimized,where $r = \max (\text{rank}(\Xb),\text{rank}(\Yb))$ and $\norm{\cdot}_{F}$ is the  Frobenius norm.

The problem has a well-known closed-form solution (see e.g., \cite{de2005eigenproblems}), relating canonical correlations and canonical pairs to the eigen-system of the following matrix
\begin{equation}\label{LCC}
\begin{bmatrix}
(\Sigmab_{xx}+\mu_x\Ib)^{-1}&{\bf 0}\\
{\bf 0}&(\Sigmab_{yy}+\mu_y\Ib)^{-1}
\end{bmatrix}
\begin{bmatrix}
{\bf 0}&\Sigmab_{xy}\\
\Sigmab_{yx}&{\bf 0}
\end{bmatrix},
\end{equation}
where $\Sigmab_{xx}=\Xb^\top\Xb,\Sigmab_{yy}=\Yb^\top\Yb,\Sigmab_{xy}=\Xb^\top\Yb$, and $\mu_x,\mu_y$ are regularization parameters to avoid singularity. 
In particular, the eigenvalues correspond to the canonical correlations and the eigenvectors correspond to the canonical pairs.

The kernel version of CCA, called KCCA \cite{lai2000kernel,bach2002kernel},  investigates the correlation analysis using the eigen-system of the following matrix
\begin{equation}\label{KCC}
\begin{bmatrix}
(\Kb_{x}+\mu_x\Ib)^{-1}&{\bf 0}\\
{\bf 0}&(\Kb_{y}+\mu_y\Ib)^{-1}
\end{bmatrix}
\begin{bmatrix}
{\bf 0}&\Kb_{y}\\
\Kb_{x}&{\bf 0}
\end{bmatrix},
\end{equation}
where $[\Kb_x]_{ij}=k_x(\xb_i,\xb_j)$ and $[\Kb_y]_{ij}=k_y(\yb_i,\yb_j)$. 

As the inversion of kernel matrices involves $O(n^3)$ time cost, \cite{lopez2014randomized} adopted the idea of kernel approximation with random features, introducing Randomized Canonical Correlation Analysis (RCCA). RCCA uses approximations $\Kb_x\approx\Zb_x\Zb_x^\top$ and  $\Kb_y\approx\Zb_y\Zb_y^\top$ in \eqref{KCC}, where $\Zb_x$ and $\Zb_y$ are the transformed matrices using random features as in \eqref{trasformed matrix}. In other words, $\text{RCCA}(\Xb,\Yb) = \text{CCA}(\Zb_x,\Zb_y) \approx \text{KCCA}(\Xb,\Yb).$ Now, the question we would like to answer in this paper is as follows:
\begin{problem}\label{P2}
If we want to maximize the total canonical correlations, i.e., the trace of matrix \eqref{KCC}, what is the corresponding score function in the form of \eqref{q} to re-weight (or select) the random features? 
\end{problem}
Note that the term ``maximize'' makes sense here due to the approximation using random features. In other words, we are interested in finding the features providing more correlation (or maximize the correlation) between $\Xb$ and $\Yb$. In the next subsection, we will derive the desired score function and show its performance advantage in theory. We will see that the features that we select based on the score function change the kernel in a way that improves the correlation.

\subsection{Related Literature}

{\bf Random features:} As discussed in Section \ref{sec:random}, kernels of form \eqref{kernel function} can be approximated using random features (e.g., shift-invariant kernels using Monte Carlo \cite{rahimi2008random} or Quasi Monte Carlo \cite{yang2014quasi} sampling, and dot product kernels \cite{kar2012random}. A number of methods have been proposed to improve the time cost, decreasing it by a linear factor of the input dimension (see e.g., Fast-food \cite{le2013fastfood,yang2015carte}). The generalization properties of random features have been studied for $\ell_1$-regularized risk minimization \cite{yen2014sparse} and ridge regression \cite{rudi2016generalization}, both improving the early generalization bound of \cite{rahimi2009weighted}. 
Also, \cite{felix2016orthogonal} develop Orthogonal Random Features (ORF) to improve kernel approximation variance. It turns out that ORF provides optimal kernel estimator in terms of mean-squared error \cite{choromanski2018geometry}. \cite{may2019kernel} present an iterative gradient method for selecting the best set of random features for supervised learning with acoustic model. A number of recent works have focused on kernel approximation techniques based on {\it data-dependent} sampling of random features. Examples include \cite{yu2015compact} on compact nonlinear feature maps, \cite{yang2015carte,oliva2016bayesian} on approximation of shift-invariant/translation-invariant kernels, \cite{chang2017data} on Stein effect in kernel approximation, and \cite{agrawal2019data} on data-dependent approximation using greedy approaches (e.g., Frank-Wolfe). On the other hand, another line of research has focused on {\it generalization} properties of data-dependent sampling. In addition to works mentioned in Section \ref{sec:relation}, \cite{bullins2017not} also study data-dependent approximation of  translation-invariant/rotation-invariant kernels for improving generalization in SVM. \cite{li2019learning} recently propose a hybrid approach (based on importance sampling) to re-weight random features with application to both kernel approximation and supervised learning.

{\bf Canonical Correlation Analysis:}
As discussed in Section \ref{sec:ccaoverview}, the computational cost of KCCA \cite{lai2000kernel} motivated a great deal of research on kernel approximation for CCA in large-scale learning. Several methods tackle this issue by explicitly transforming datasets (e.g., Randomized Canonical Correlation Analysis (RCCA) \cite{lopez2013randomized,lopez2014randomized}, Fix-Sized KernelCCA (FSCCA) \cite{mehrkanoon2017regularized}, and Deep Canonical Correlation Analysis (DCCA) \cite{andrew2013deep}). RCCA and FSCCA tackle the computation issue of KCCA with two well-known kernel approximation methods, random features and Nystrom method respectively. RCCA focuses on transformation using randomized 1-hidden layer neural networks, whereas DCCA considers deep neural networks. Perhaps not surprisingly, the time cost of RCCA is significantly smaller than DCCA \cite{lopez2014randomized}. There exists other non-parametric approaches such as Non-parametric Canonical Correlation Analysis (NCCA) \cite{michaeli2016nonparametric}, which estimates the density of training data to provide a practical solution to Lancaster’s theory for CCA \cite{lancaster1958structure}. Also, more recently, a method is proposed in \cite{uurtio2019large} for sparsifying KCCA through $\ell_1$ regularization. A different (but relevant) literature has focused on addressing the optimization problem in CCA. \cite{wang2016efficient,arora2017stochastic} have discussed this problem by developing novel techniques, such as alternating least
squares, shift-and-invert preconditioning, and inexact matrix stochastic gradient. In a similar spirit is \cite{wang2016large}, which presents a memory-efficient stochastic optimization algorithm for RCCA. 

The main novelty of our approach is proposing an optimized scoring rule for random features selection, which can be adopted for different learning tasks including various correlation analysis techniques, e.g, CCA. 

\section{Objective Based Score Function}\label{sec:generalscore}

\subsection{A Task-Specific Scoring Rule for Random Features Selection}\label{sec:score}
Several recent works have answered to Problem \ref{P1} in the affirmative; however, quite interestingly, there is so much difference in adopted strategies given the learning task. For example, a sampling scheme that improves kernel approximation (e.g., Orthogonal Random Features \cite{felix2016orthogonal}) will not necessarily be competitive for supervised learning \cite{shahrampour2018data}. In other words, Problem \ref{P1} has been addressed in a {\it task-specific} fashion.   
In this paper, we propose a scoring rule for selecting features that lends itself to several important tasks in machine learning. Let $\Bb$ be a real matrix and define the following score function for any $\omegab \in \Omega$
\begin{align}\label{q}
    q(\omegab) \triangleq p(\omegab)\zb^\top(\omegab)\Bb~\zb(\omegab),
\end{align}
where $p(\omegab)$ is the original probability density of random features. $p(\omegab)$ can be thought as an easy prior to sample from. The score function $q(\omegab)$ can then serve as the metric to re-weight the random features from prior $p(\omegab)$. The key advantage of the score function is that $\Bb$ can be selected based on the learning task to improve the performance. We will elaborate on this choice in Subsections \ref{sec:relation}-\ref{sec:LinktoDR}.

\subsection{Relation to Supervised Learning Scoring Rules}\label{sec:relation}
A number of recent works have proposed the idea of sampling random features based on {\it data-dependent} distributions, mostly focusing on improving generalization in supervised learning. In this section, we show that the score function \eqref{q} will bring some of these methods under the same umbrella. More specifically, given a particular choice of the center matrix $\Bb$, we can recover a number of data-dependent sampling schemes, such as Leverage Scores (LS) \cite{avron2017random,bach2017equivalence,rudi2016generalization,sun2018but} and Energy-based Exploration of Random Features (EERF) \cite{shahrampour2018data}. 

{\bf Leverage Scores:} Following the framework of \cite{avron2017random}, LS sampling is according to the following probability density function
\begin{equation}\label{qelss}
    q_{LS}(\omegab) \propto p(\omegab)\zb^\top(\omegab)(\Kb+\lambda\Ib)^{-1}\zb(\omegab),
\end{equation}
which can be recovered precisely when $\Bb=(\Kb+\lambda\Ib)^{-1}$ in \eqref{q}. A practical implementation of LS was proposed in \cite{bach2017equivalence} and later used in the experiments of \cite{sun2018but} for SVM. The generalization properties of (a variant of) LS algorithm was also studied by \cite{li2019towards} for the case of ridge regression.

{\bf Energy-Based Exploration of Random Features:} The EERF algorithm was proposed in \cite{shahrampour2018data} for improving generalization. In supervised learning, the goal is to map input vectors $\{\xb_i\}_{i=1}^n$ to output variables $\{y_i\}_{i=1}^n$, where $y_i\in \R$ for $i\in [n]$. The EERF algorithm employs the following scoring rule for random features
\begin{equation}\label{qeerf}
    q_{\text{EERF}}(\omegab) \propto
 \abs{\frac{1}{n}\sum_{i=1}^n y_i\phi(\xb_i,\omegab)},
\end{equation}
where the score is calculated for a large pool of random features, and a subset with the largest score will be used for the supervised learning problem. Now, if we let $\yb = [y_1,\ldots,y_n]^\top$, we can observe that $q_{\text{EERF}}(\omegab)$ is equivalent to \eqref{q} with the center matrix $\Bb=\yb\yb^\top$, because ordering the pool of features according to $(\yb^\top\zb(\omegab))^2=(\sum_{i=1}^n y_i\phi(\xb_i,\omegab))^2$ is equivalent to $\abs{\frac{1}{n}\sum_{i=1}^n y_i\phi(\xb_i,\omegab)}$ given above. The authors of \cite{shahrampour2018data} showed in their numerical experiments that EERF consistently outperforms plain random features and other data-independent methods in terms of generalization. We remark that the kernel alignment method in \cite{sinha2016learning} is also in a similar spirit. Instead of choosing features with largest scores, an optimization algorithm is proposed to re-weight the features such that the transformed input is correlated enough with output variable.

Given the success of algorithms like LS and EERF, we can hope that the scoring rule \eqref{q} has the potential to be adopted in various learning tasks. Indeed, the center matrix $\Bb$ should be chosen based on the objective function that needs to be optimized in the learning task at hand.

\subsection{Relation to Dimension Reduction \& Correlation Analysis}\label{sec:LinktoDR}
We now show another potential of the score function \eqref{q} by establishing that
\begin{equation}\label{trace}
\begin{aligned}
\tr{\Kb \Bb}=\int_\Omega q(\omegab) d\omegab &= \int_\Omega p(\omegab)\zb^\top(\omegab)\Bb~\zb(\omegab)\\
&\approx \frac{1}{M}\sum_{m=1}^M\zb^\top(\omegab_m)\Bb~\zb(\omegab_m),
\end{aligned}
\end{equation}
where $\{\omegab_m\}_{m=1}^M$ are independent samples from $p(\omegab)$. The above relationship reveals the connection of the score function with a class of trace maximization problems dealing with kernelized objective functions. Several dimension reduction/correlation analysis methods fall into this category. For example, Kernel Principal Component Analysis (KPCA) and Kernel Orthogonal Neighborhood Preserving Projections (KONPP) have the exact form of the objective function in \eqref{trace} (see e.g.,  \cite{kokiopoulou2011trace}). In each case, we can identify the matrix $\Bb$ by looking at the corresponding eigenvalue problem. For KPCA, the eigenvalue problem implies that the matrix $\Bb=(\Ib - \frac{1}{n}{\bf 1}{\bf 1}^\top)$, where ${\bf 1}$ is the vector of all ones. In KONPP, the corresponding eigenvalue problem entails that $\Bb=(\Ib - \Wb^\top)(\Ib - \Wb)$, where $\Wb$ is the affinity matrix in the feature space. Now, instead of covering more high level formulations, in the next section, we will carefully study Canonical Correlation Analysis, which is also formulated as a trace optimization \eqref{trace}.

\begin{remark}
The scoring rule \eqref{q} offers a principled way to select random features that promise good performance for a specific objective. In this section, we identify a class of dimension reduction and correlation analysis problems that is compatible with the proposed scoring rule. However, its adaptation to other problems still requires efforts in the identification and derivation of the score function. In Section \ref{sec:CCA}, we will focus on nonlinear CCA, as an important problem in machine learning and statistics, to derive the respective score function and use it for algorithm implementation.
\end{remark}

\section{Canonical Correlation Analysis with Score-Based Random Features Selection}\label{sec:CCA}
We now show the application of the scoring rule \eqref{q} to nonlinear Canonical Correlation Analysis (CCA).

\subsection{Optimal Randomized Canonical Correlation Analysis (ORCCA)}
We now propose the adaptations of the scoring rule \eqref{q} for CCA, where the center matrix $\Bb$ is selected particularly for maximizing the total canonical correlation. We start with an important special case of $d_y=1$ due to the natural connection to supervised learning. 
We will use index $x$ for any quantity in relation to $\Xb$, and $y$ for any quantity in relation to $\Yb$. {\color{red}}

{\bf Optimal Randomized Canonical Correlation Analysis 1 ($\mathbf{d_y=1}$ and linear $\Kb_y$):} We consider the scenario where $\Xb\in \R^{n\times d_x}$ is mapped into a nonlinear space $\Zb_x \in \R^{n\times M}$ (using random features) following \eqref{trasformed matrix}. On the other hand, $\Yb=\yb\in \R^n$ remains in its original space (with $d_y=1$ and $\Kb_y = \yb\yb^\top$). It is well-known that if $\yb=\Zb_x\alphab$ for some $\alphab \in \R^M$, perfect (linear) correlation is achieved between $\yb$ and $\Zb_x$
(with $\mu_x=\mu_y=0$ and $n>d_x$), simply because $\yb$ is a linear combination of the columns of $\Zb_x$. This motivates the idea that sampling schemes that are good for supervised learning may be natural candidates for CCA in that with $\yb=\Zb_x\alphab$ we can achieve perfect correlation. The following proposition finds the optimal scoring rule of form \eqref{q} that maximizes the total canonical correlation.

\begin{proposition}\label{P: PORCCA}
Consider KCCA in \eqref{KCC} with $\mu_x=\mu_y=\mu$, a nonlinear kernel matrix $\Kb_x$ and a linear kernel $\Kb_y=\yb\yb^\top$. If we approximate $\Kb_x\approx \Zb_x\Zb_x^\top$ only in the right block matrix of \eqref{KCC}, the optimal scoring rule maximizing the total canonical correlation can be expressed as
\begin{equation}\label{PORCCA}
    q(\omegab) = p(\omegab)\zb_x^\top(\omegab)(\Kb_x+\mu\Ib)^{-1}\yb\yb^\top\zb_x(\omegab),
\end{equation}
for any $\omegab\in \Omega_x \subseteq \R^{d_x}$. The scoring rule above corresponds to \eqref{q} with $\Bb=(\Kb_x+\mu\Ib)^{-1}\yb\yb^\top.$
\end{proposition}
Interestingly, the principled way of choosing $\Bb$ that maximizes total CCA leads to a feature selection rule that was not previously investigated. It is clear that the score function in \eqref{PORCCA} is different from LS \eqref{qelss} and EERF \eqref{qeerf}. While the scoring rule \eqref{PORCCA} optimizes canonical correlations in view of Proposition \ref{P1}, calculating $\Bb$ would cost $O(n^3)$, which is not scalable to large datasets. The following corollary offers an approximated solution to avoid this issue.
\begin{corollary}\label{C: PORCCA}
For any finite pool of random features $\{\omegab_m\}_{m=1}^{M_0}$, instead of selecting according to the scoring rule \eqref{PORCCA}, we can approximate the scoring rule with the following empirical score
\begin{equation}\label{emp1}
    q(\omegab_i)\approx\qh(\omegab_i) \triangleq \left[(\Zb_x^\top\Zb_x+ \mu\Ib)^{-1}\Zb_x^\top\yb\yb^\top\Zb_x \right]_{ii},
\end{equation}
for any $\omegab_{x,i}\in \Omega_x \subseteq \R^{d_x}$ and $i\in [M_0]$, where $\Zb_x$ is formed with $M_0$ random features as in \eqref{trasformed matrix} and $\qh(\omegab_i)$ denotes the empirical score of the $i$-th random features in the pool of $M_0$ features.
\end{corollary}
Observe that selecting according to the score rule above will reduce the computational cost from $O(n^3)$ to $O(nM_0^2+M_0^3)$, which is a significant improvement when $M_0 \ll n$. After constructing \eqref{emp1}, we can select the $M$ features with highest empirical scores. This algorithm is called ORCCA1 presented in Algorithm \ref{ALGO1}.

\begin{algorithm}[ht]
	\caption{Optimal Randomized Canonical Correlation Analysis 1 (ORCCA1)}
	{\bf Input:} 
	$\Xb \in \R^{n\times d_x}$,$\yb\in \R^n$, the feature map $\phi(\cdot,\cdot)$, an integer $M_0$, an integer $M$, the prior distribution $p(\omegab)$, the parameter $\mu>0$.
	\begin{algorithmic}[1]
	\STATE Draw $M_0$ independent samples $\{\omegab_m\}_{m=1}^{M_0}$ from $p(\omegab)$.
	\STATE Construct the matrix 
    \begin{equation*}
        \Qb=(\Zb_x^\top\Zb_x+ \mu\Ib)^{-1}\Zb_x^\top\yb\yb^\top\Zb_x,
    \end{equation*}
	where $\Zb_x$ is defined in \eqref{trasformed matrix}.
	\STATE Let for $i\in [M_0]$
	\begin{align*}
	    \qh(\omegab_i)=[\Qb]_{ii}.
	\end{align*}
	The new weights $\qbh=[\qh(\omegab_1),\ldots,\qh(\omegab_{M_0})]^\top$.
	\STATE Sort $\qbh$ and select top $M$ features with highest scores from the pool to construct the transformed matrix $\widehat{\Zb}_x$ following \eqref{trasformed matrix}.
	\end{algorithmic}
	{\bf Output:} 
	Linear canonical correlations between $\widehat{\Zb}_x$ and $\yb$ (with regularization parameter $\mu$). 
\label{ALGO1}
\end{algorithm}

{\bf Optimal Randomized Canonical Correlation Analysis 2 (nonlinear $\Kb_y$):} We now follow the idea of KCCA with both views of data mapped to a nonlinear space. More specifically, $\Xb\in \R^{n\times d_x}$ is mapped to $\Zb_x \in \R^{n\times M}$ and $\Yb\in \R^{n\times d_y}$ is mapped to $\Zb_y \in \R^{n\times M}$ following \eqref{trasformed matrix}. For this set up, we provide below the optimal scoring rule of form \eqref{q} that maximizes the total canonical correlation. 

\begin{theorem}\label{P: DORCCA}
Consider KCCA in \eqref{KCC} with $\mu_x=\mu_y=\mu$, a nonlinear kernel matrix $\Kb_x$, and a nonlinear kernel $\Kb_y$. If we alternatively approximate $\Kb_x\approx \Zb_x\Zb_x^\top$ and $\Kb_y\approx \Zb_y\Zb_y^\top$ only in the right block matrix of \eqref{KCC}, the optimal scoring rule maximizing the total canonical correlation can be expressed as
\begin{equation}\label{DORCCA}
\begin{aligned}
    q_x(\omegab) &= p_x(\omegab)\zb^\top_x(\omegab)(\Kb_x+\mu\Ib)^{-1}\Kb_y(\Kb_y+\mu\Ib)^{-1}\zb_x(\omegab)\\
    q_y(\omegab') &= p_y(\omegab')\zb^\top_y(\omegab')(\Kb_y+\mu\Ib)^{-1}\Kb_x(\Kb_x+\mu\Ib)^{-1}\zb_y(\omegab'),
\end{aligned}
\end{equation}
for any $\omegab\in \Omega_x \subseteq \R^{d_x}$ and any $\omegab'\in \Omega_y \subseteq \R^{d_y}$, respectively. The probability densities $p_x(\omegab)$ and $p_y(\omegab')$ are the priors defining the default kernel functions in the space of $\Xc$ and $\Yc$ according to \eqref{kernel function}. \end{theorem}
We can associate the scoring rules above to the task-specific scoring rule \eqref{q} as well. Indeed, for choosing the random features from $\Omega_x$ to transform $\Xb$, the center matrix is $\Bb=(\Kb_x+\mu\Ib)^{-1}\Kb_y(\Kb_y+\mu\Ib)^{-1}$, and for choosing the random features from $\Omega_y$ to transform $\Yb$, the center matrix is $\Bb=(\Kb_y+\mu\Ib)^{-1}\Kb_x(\Kb_x+\mu\Ib)^{-1}$. While the scoring rule \eqref{DORCCA} optimizes canonical correlations in view of \eqref{KCC}, calculating $\Bb$ would cost $O(n^3)$, which is not scalable to large datasets. The following corollary offers an approximated solution to avoid this issue.

\begin{corollary}\label{C: DORCCA}
For any finite pool of random features $\{\omegab_{x,m}\}_{m=1}^{M_0}$ and $\{\omegab_{y,m}\}_{m=1}^{M_0}$ (sampled from priors $p_x(\omegab)$ and $p_y(\omegab)$, respectively), 
instead of selecting according to the scoring rules \eqref{DORCCA}, we can approximate them using the following empirical versions
\begin{align*}
    \hat{q}_x(\omegab_{x,i})&= \left[(\Zb_x^\top\Zb_x + \mu\Ib)^{-1}\Zb_x^\top\Zb_y(\Zb_y^\top\Zb_y + \mu\Ib)^{-1}\Zb_y^\top\Zb_x\right]_{ii}\\
    \hat{q}_y(\omegab_{y,i})&= \left[(\Zb_y^\top\Zb_y + \mu\Ib)^{-1}\Zb_y^\top\Zb_x(\Zb_x^\top\Zb_x + \mu\Ib)^{-1}\Zb_x^\top\Zb_y\right]_{ii},
\end{align*}
for any $\omegab_{x,i}\in \Omega_x \subseteq \R^{d_x}$ and any $\omegab_{y,i}\in \Omega_y \subseteq \R^{d_y}$, respectively. $\Zb_x$ and $\Zb_y$ are the transformed matrices of $\Xb$ and $\Yb$ as in \eqref{trasformed matrix} using $M_0$ random features. $\qh_x(\omegab_{x,i})$ and $\qh_y(\omegab_{y,i})$ denote the scores of the $i$-th random features in the pools corresponding to $\Xb$ and $\Yb$, respectively. 
\end{corollary}
As we observe, the computational cost in both view of the data is reduced from $O(n^3)$ to $O(nM_0^2+M_0^3)$. The justification is provided in the appendix (Subsection \ref{comp}). We use the above empirical scores for the implementation of ORCCA2, described in Algorithm \ref{ALGO2}. We also prove below that the theoretical score functions in \eqref{PORCCA} and \eqref{DORCCA} always provide improvement over RCCA and KCCA.

\begin{proposition}\label{upperbound}
The total canonical correlation obtained with $M$ features selected from an $M_0$ features pool ($M < M_0 < \infty$) using non-empirical scores \eqref{PORCCA} and \eqref{DORCCA} provides an theoretical upper bound of the total canonical correlation obtained with $M$ plain random Fourier features in expectation. 
Let us denote by $\rho(\text{KCCA})$ the total canonical correlation obtained by KCCA and by $\rho^{(M)}(\text{RCCA})$ the total canonical correlation obtained by RCCA with $M$ plain random Fourier features. Let us also represent by $\rho^{(M,M_0)}(\text{ORCCA})$ the total canonical correlation obtained by ORCCA (1 and 2), where top $M$ features are selected from a pool of $M_0>M$ plain random Fourier features according to score \eqref{DORCCA}. Then, the following relationship holds
\begin{equation}
    \E\bigg[ \rho^{(M,M_0)}(\text{ORCCA}) \bigg] \geq \E \bigg[ \rho^{(M)}(\text{RCCA}) \bigg]=\rho(\text{KCCA}),
\end{equation}
where the expectation is taken over random features.
\end{proposition}
The intuition is that the selected features via the proposed score obtain a sup value over the subsets of the feature pool, therefore, the expectation of the sup will be greater than the sup of expectation. This logic applies to other adaptations as well, meaning that proper adaptation of score function \eqref{q} will provide uniformly better results than random Fourier features in any suitable tasks. 
The proof of our results are given in the appendix. 

\begin{algorithm}[ht]
	\caption{Optimal Randomized Canonical Correlation Analysis 2 (ORCCA2)}
	{\bf Input:} 
	$\Xb \in \R^{n\times d_x}$,$\Yb\in \R^{n\times d_y}$, the feature map $\phi(\cdot,\cdot)$, an integer $M_0$, an integer $M$, the prior densities $p_x(\omegab)$ and $p_y(\omegab)$, parameter $\mu>0$.
	\begin{algorithmic}[1]
	\STATE Draw samples $\{\omegab_{x,m}\}_{m=1}^{M_0}$ and $\{\omegab_{y,m}\}_{m=1}^{M_0}$ according to $p_x(\omegab)$ and $p_y(\omegab)$, respectively. 
	\STATE Construct the matrices
    \begin{align*}
    \Qb&=(\Zb_x^\top\Zb_x+ \mu\Ib)^{-1}\Zb_x^\top\Zb_y\\
    \Pb&=(\Zb_y^\top\Zb_y+ \mu\Ib)^{-1}\Zb_y^\top\Zb_x.
    \end{align*}
    where $\Zb_x$ and $\Zb_y$ are defined in \eqref{trasformed matrix}.
	\STATE Let for $i\in [M_0]$
	\begin{align*}
	    \qh_x(\omegab_{x,i})=[\Qb\Pb]_{ii}.
	\end{align*}
	The new weights $\qbh_x=[\qh_x(\omegab_{x,1}),\ldots,\qh_x(\omegab_{x,M_0})]^\top$.
	\STATE Let for $i\in [M_0]$
	\begin{align*}
	    \qh_y(\omegab_{y,i})=[\Pb\Qb]_{ii}.
	\end{align*}
	The new weights $\qbh_y=[\qh_y(\omegab_{y,1}),\ldots,\qh_y(\omegab_{y,M_0})]^\top$.
	\STATE Select top $M$ features with the highest scores from each of the pools $\{\omegab_{x,i}\}_{i=1}^{M_0}$ and $\{\omegab_{y,i}\}_{i=1}^{M_0}$, according to the new scores $\qbh_x$ and $\qbh_y$ to construct the transformed matrices $\widehat{\Zb}_x \in \R^{n\times M}$ and $\widehat{\Zb}_y \in \R^{n\times M}$, respectively, as in \eqref{trasformed matrix}.
	\end{algorithmic}
	{\bf Output:} 
	Linear canonical correlations between $\widehat{\Zb}_x$ and $\widehat{\Zb}_y$ (with parameter $\mu$).
\label{ALGO2}
\end{algorithm}

\begin{remark}
Notice ORCCA1 is a special case of ORCCA2 where $d_y = 1$ and $\Kb_y = \yb\yb^\top$, we are presenting it as a separate algorithm to highlight its connection with supervised learning.
\end{remark}

\begin{figure*}[t!]
\centering
\includegraphics[width = 0.24\textwidth, height=0.15\textheight]{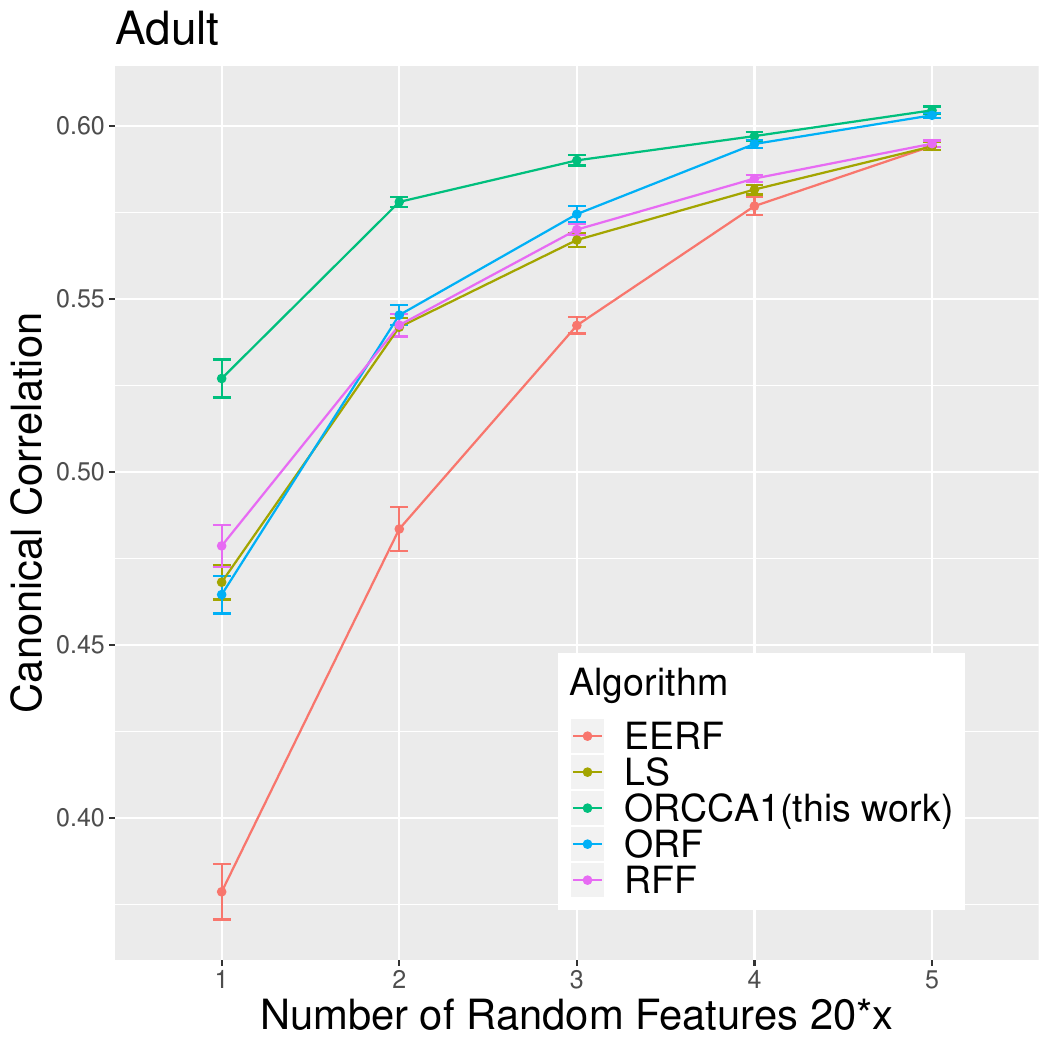}
\includegraphics[width = 0.24\textwidth, height=0.15\textheight]{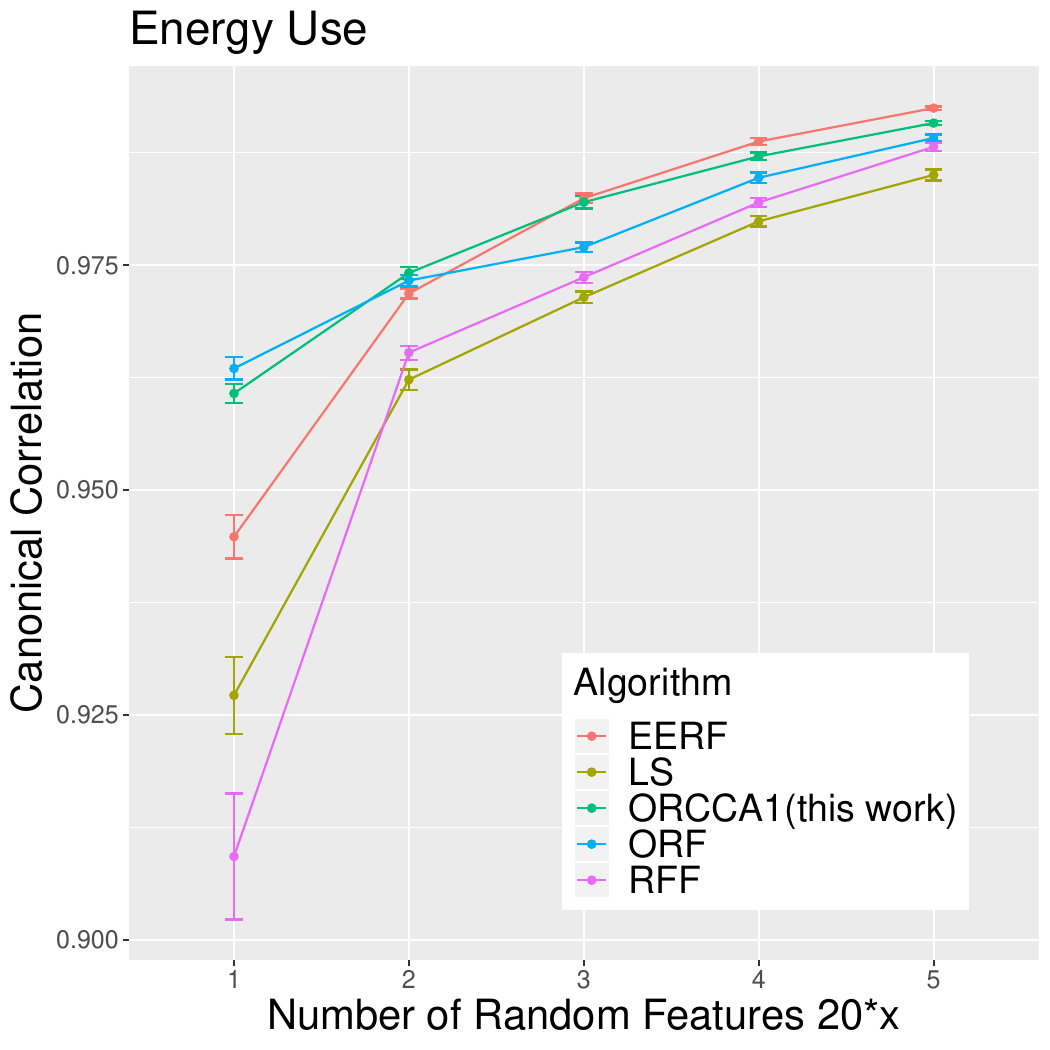}
\includegraphics[width = 0.24\textwidth, height=0.15\textheight]{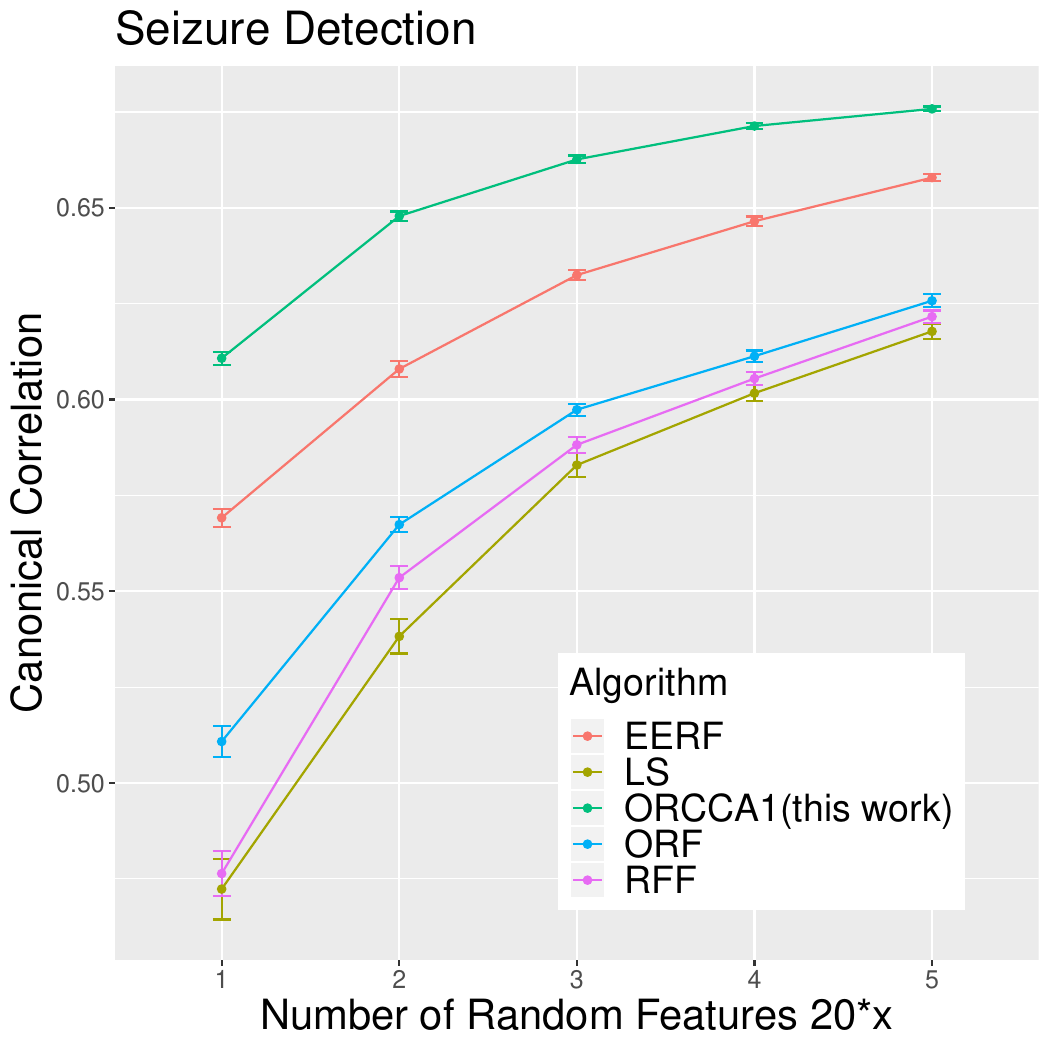}
\includegraphics[width = 0.24\textwidth, height=0.15\textheight]{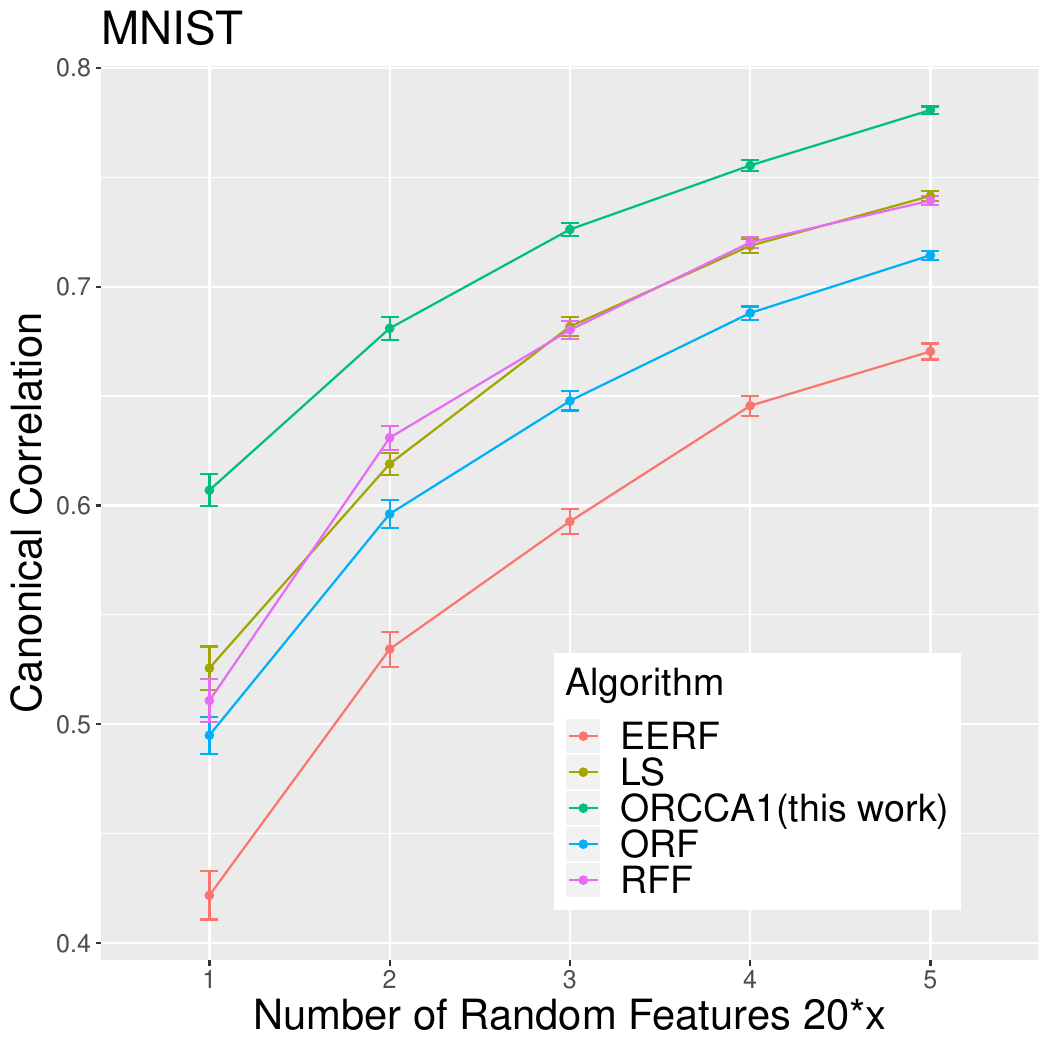}
\caption{The plot of canonical correlations versus the number of features obtained by different algorithms (for ORCCA1 comparison). The error bars are obtained with $30$ Monte-Carlo simulations.}
\label{fig: ORCCA1}
\end{figure*}

\begin{figure*}[t!]
\centering
\includegraphics[width = 0.24\textwidth, height=0.13\textheight]{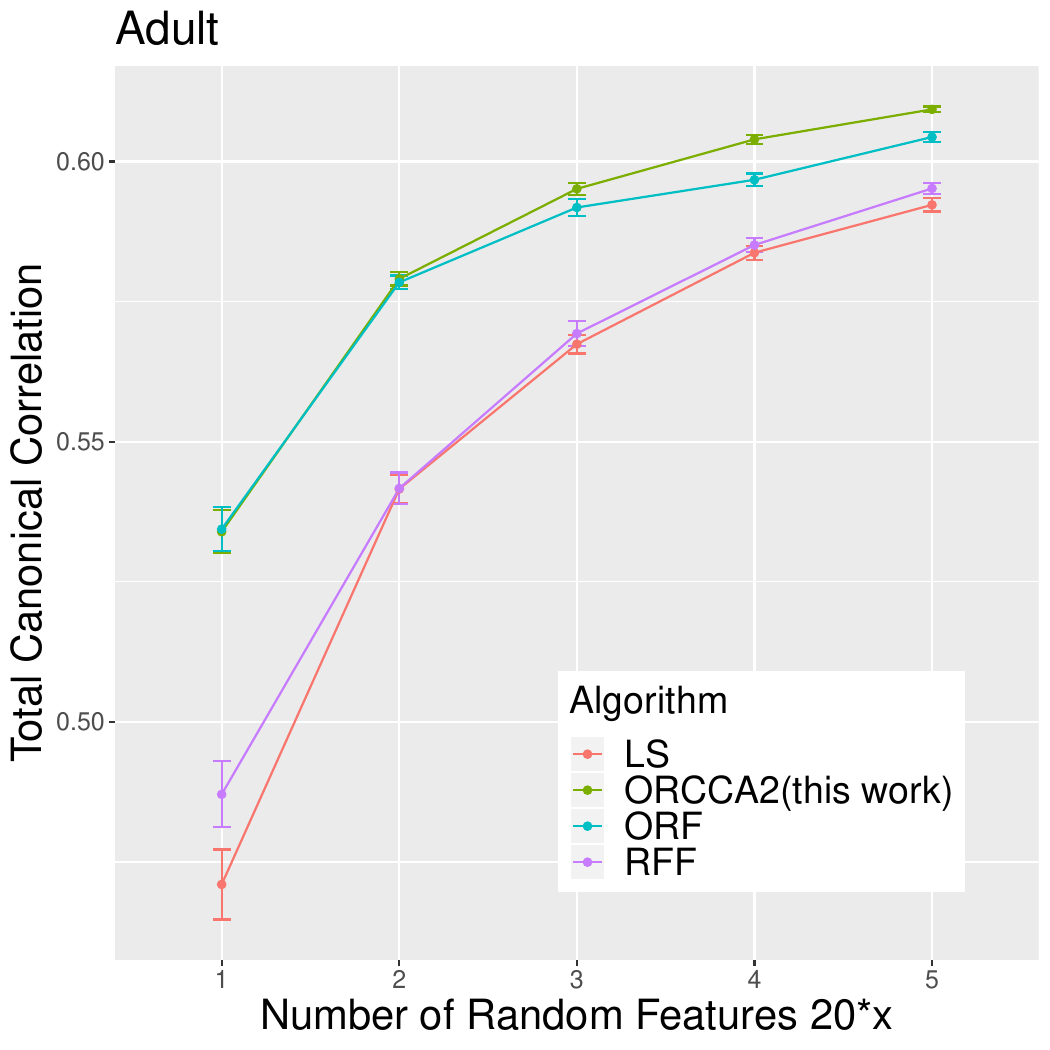}
\includegraphics[width = 0.24\textwidth, height=0.13\textheight]{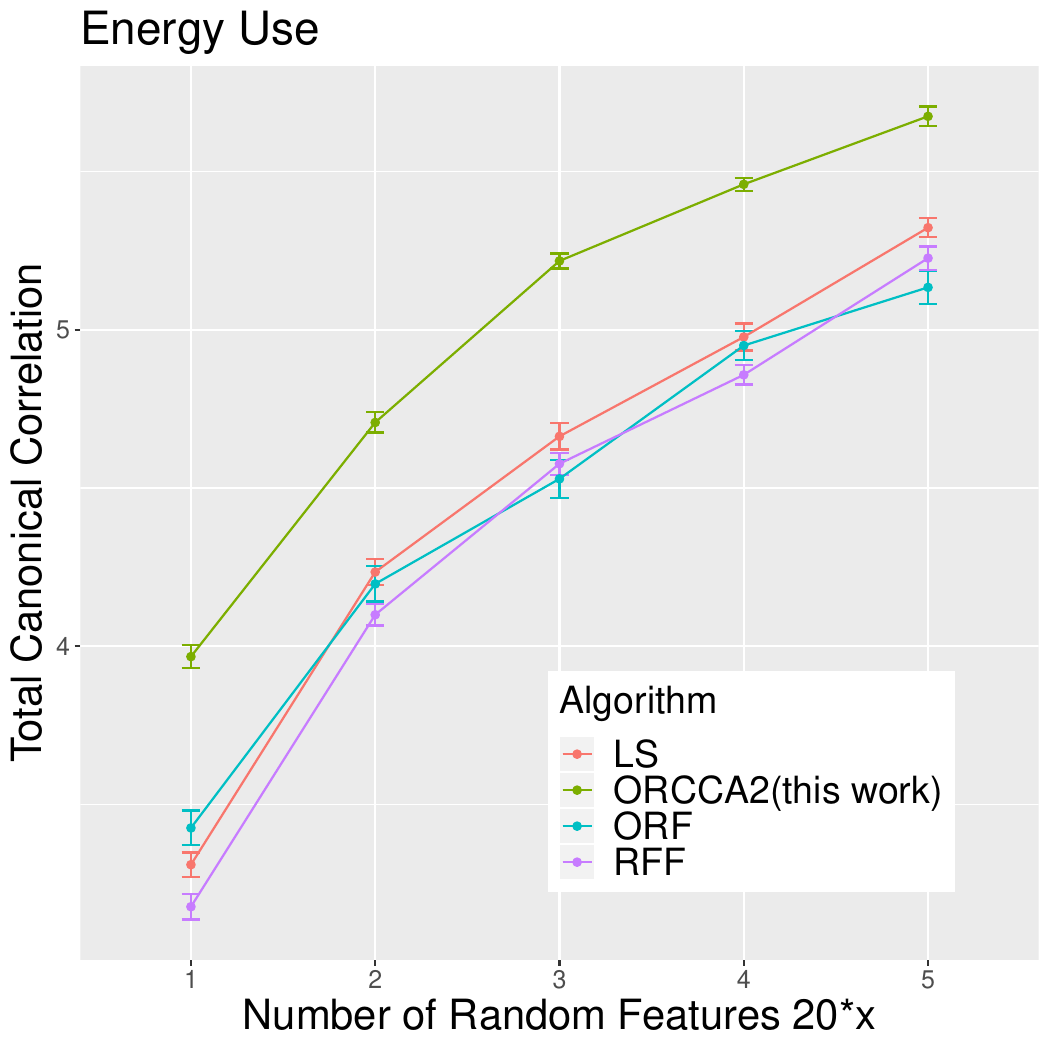}
\includegraphics[width = 0.24\textwidth, height=0.13\textheight]{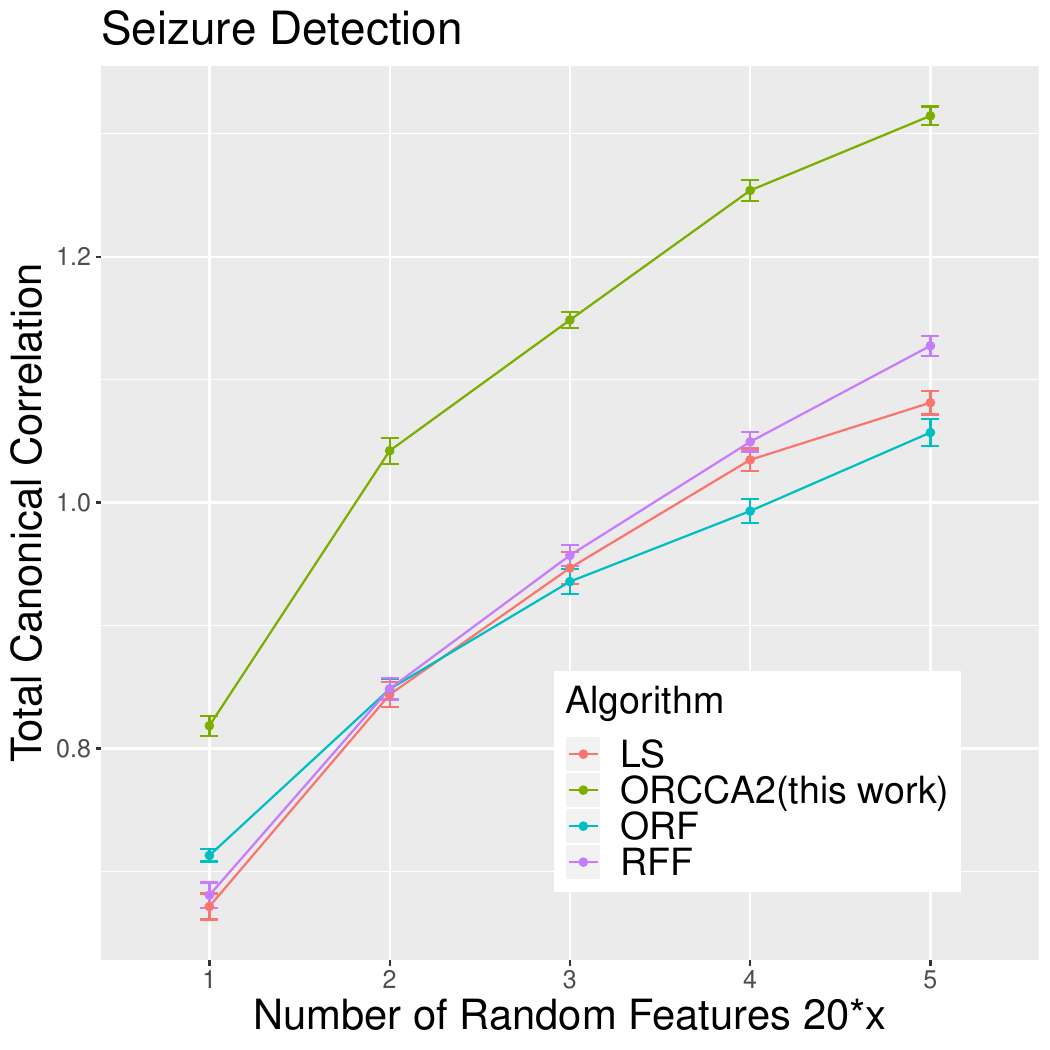}
\includegraphics[width = 0.24\textwidth, height=0.13\textheight]{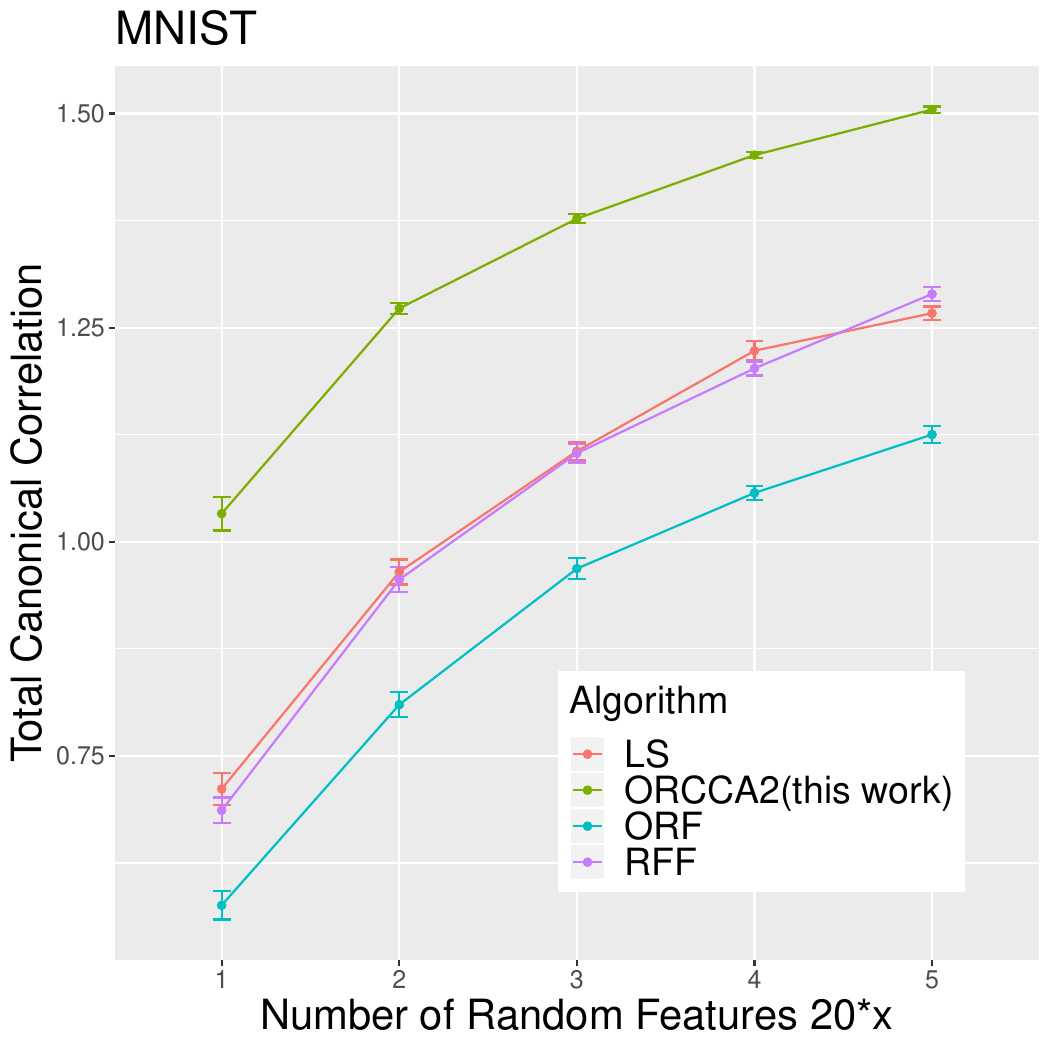}\\
\includegraphics[width = 0.24\textwidth, height=0.13\textheight]{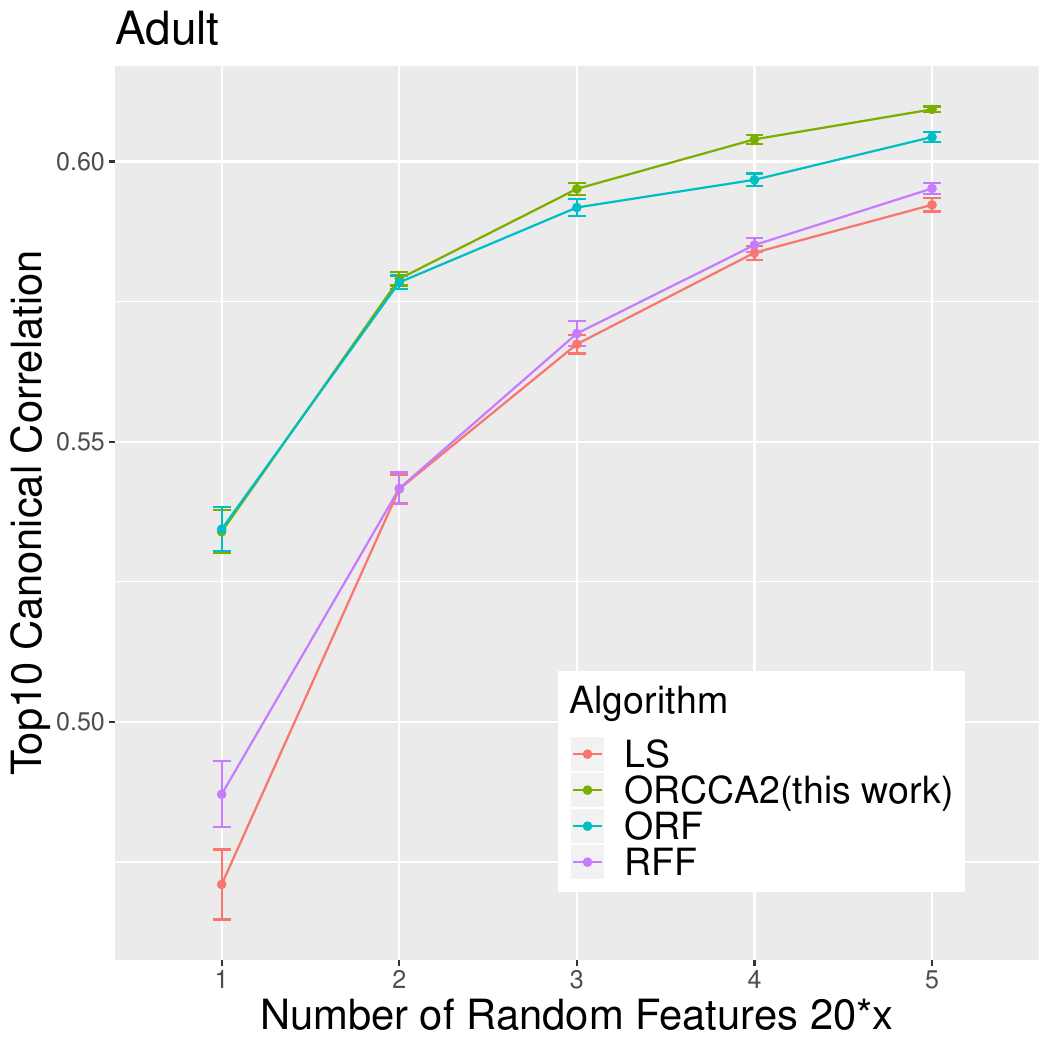}
\includegraphics[width = 0.24\textwidth, height=0.13\textheight]{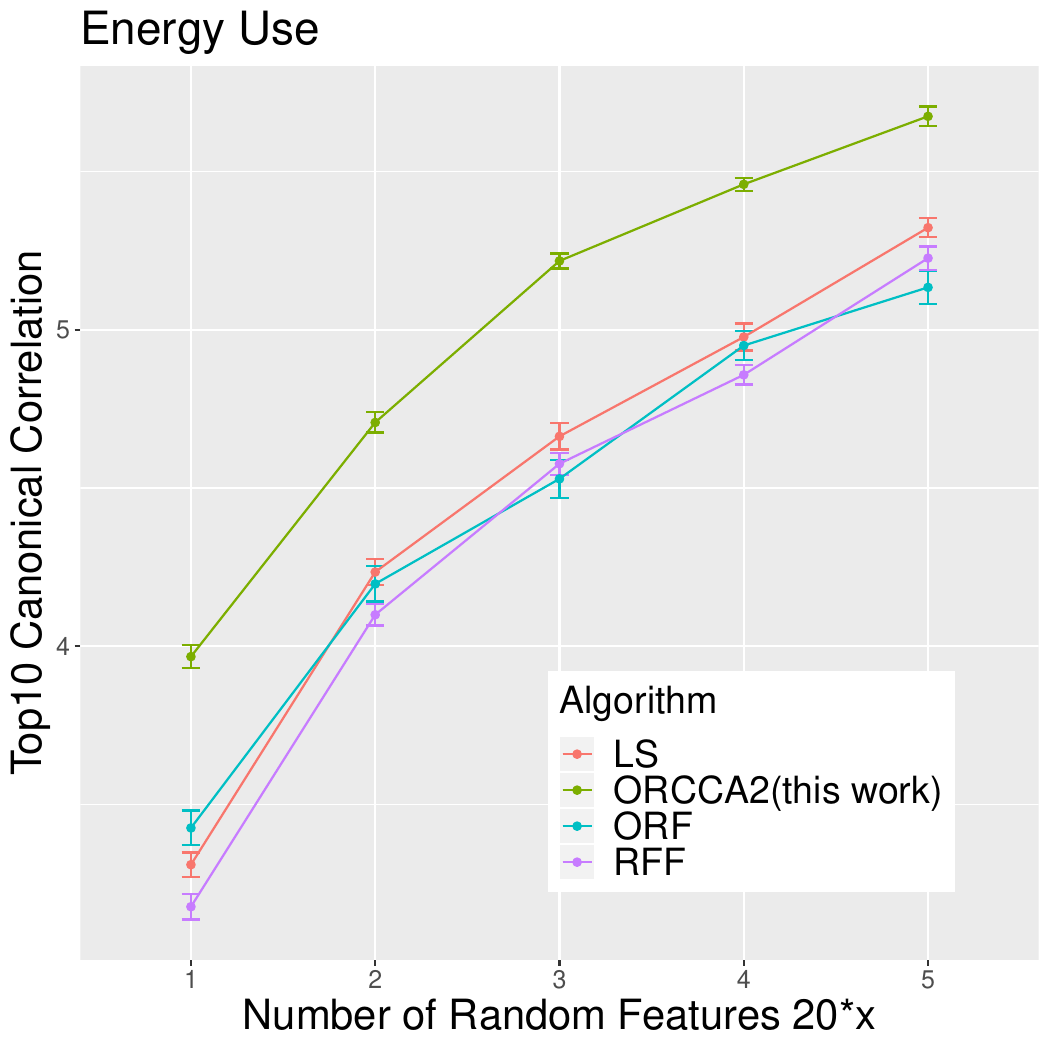}
\includegraphics[width = 0.24\textwidth, height=0.13\textheight]{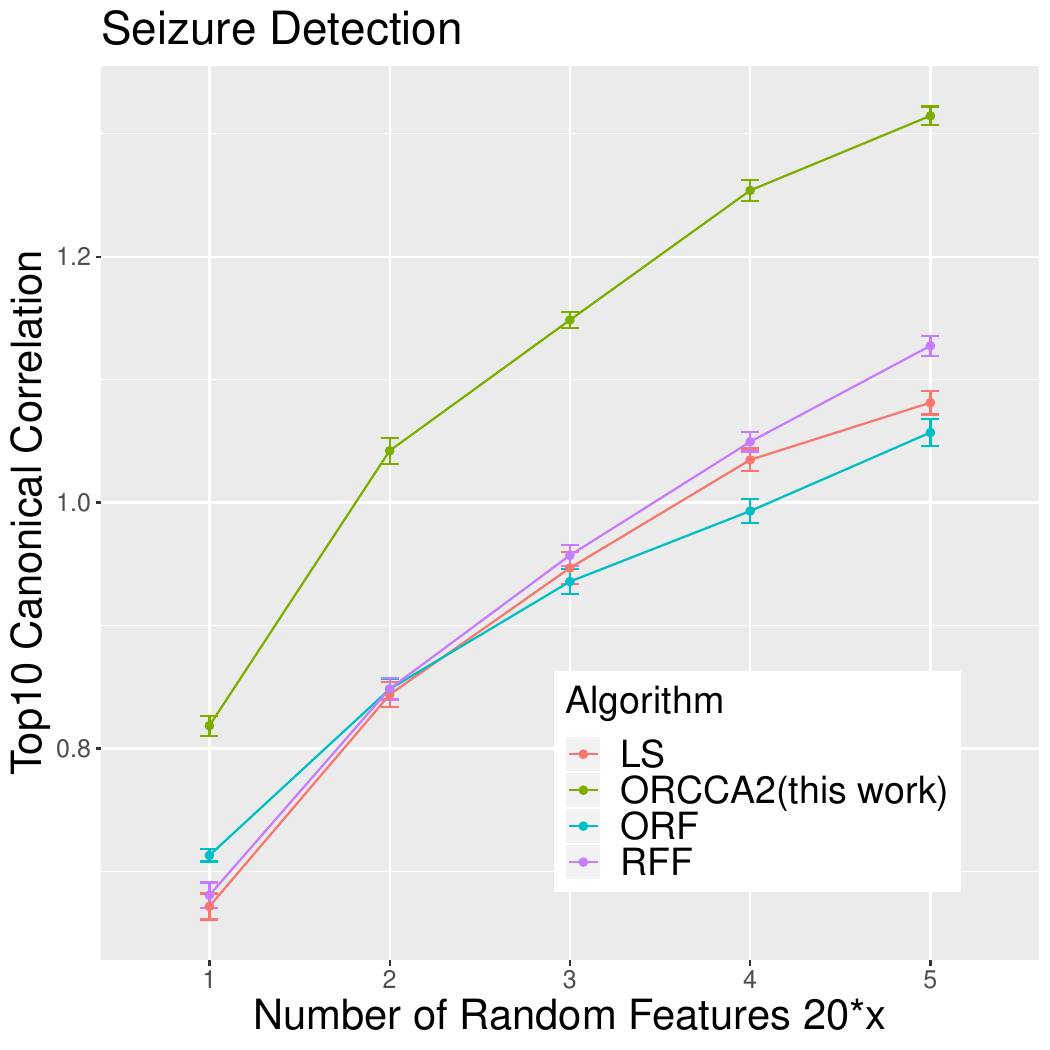}
\includegraphics[width = 0.24\textwidth, height=0.13\textheight]{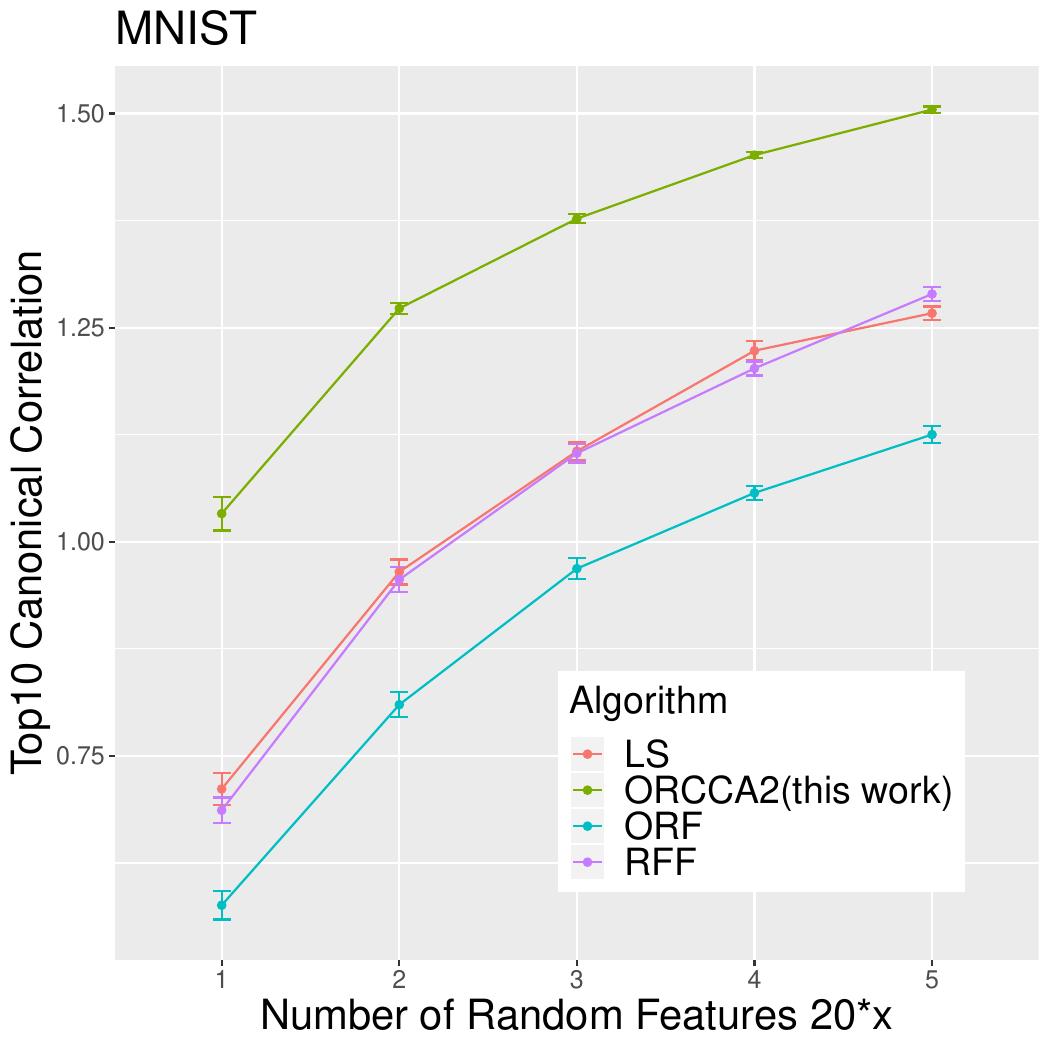}\\
\includegraphics[width = 0.24\textwidth, height=0.13\textheight]{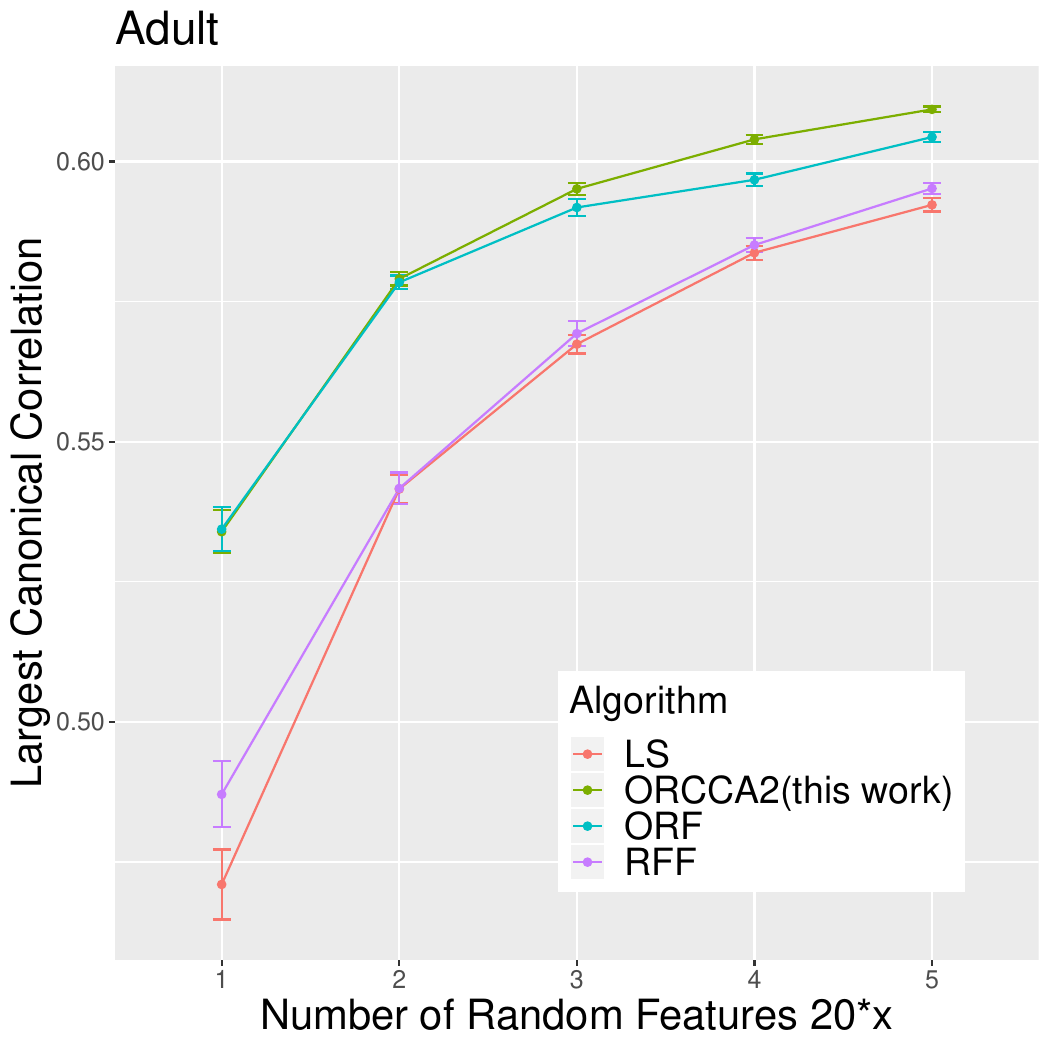}
\includegraphics[width = 0.24\textwidth, height=0.13\textheight]{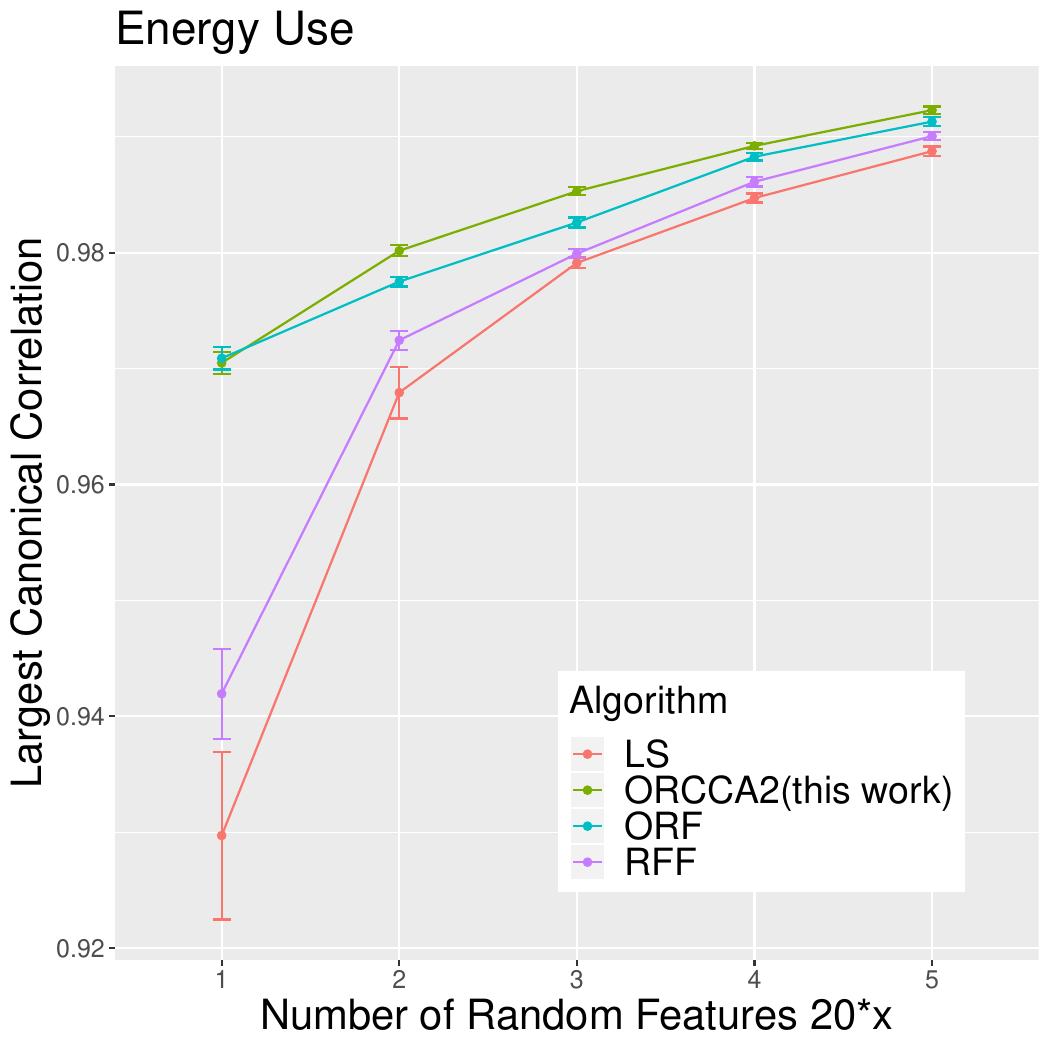}
\includegraphics[width = 0.24\textwidth, height=0.13\textheight]{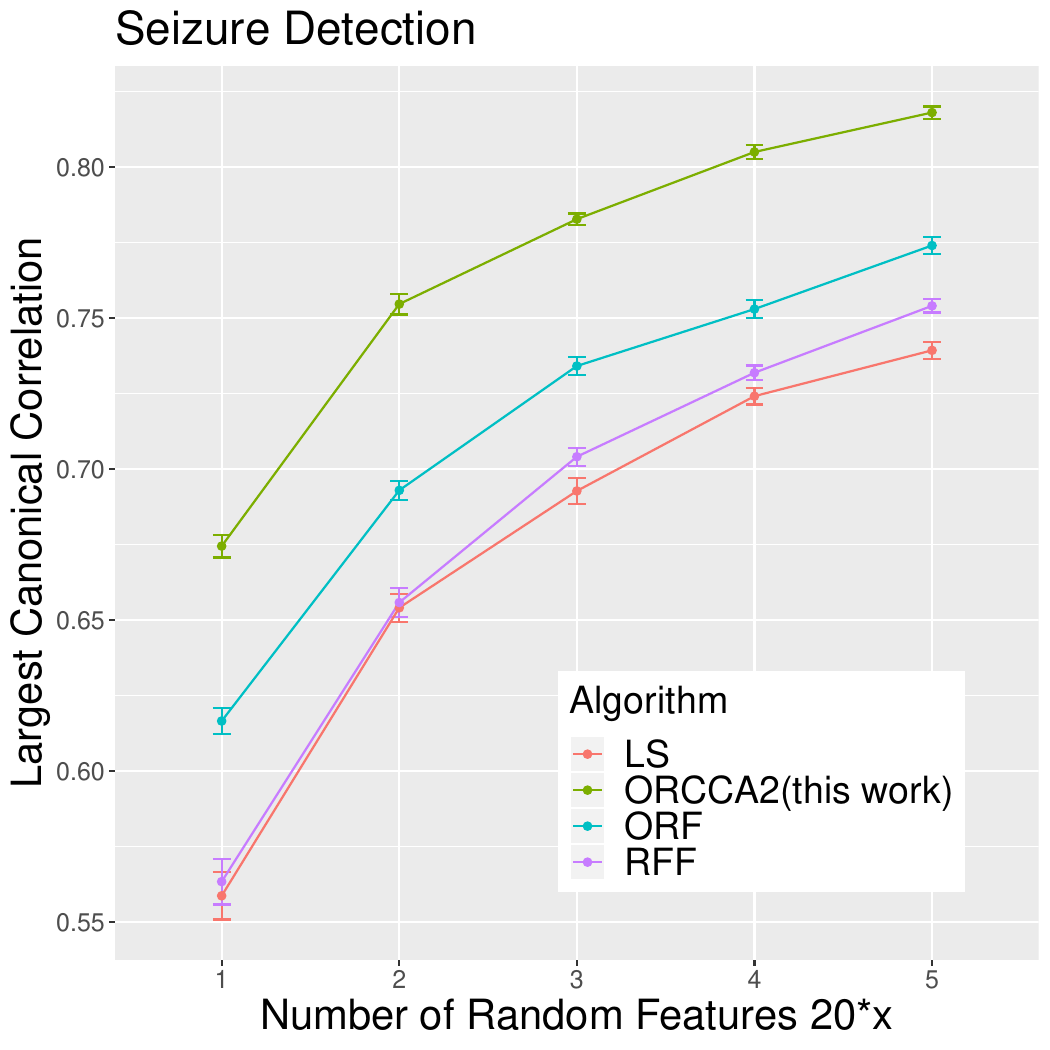}
\includegraphics[width = 0.24\textwidth, height=0.13\textheight]{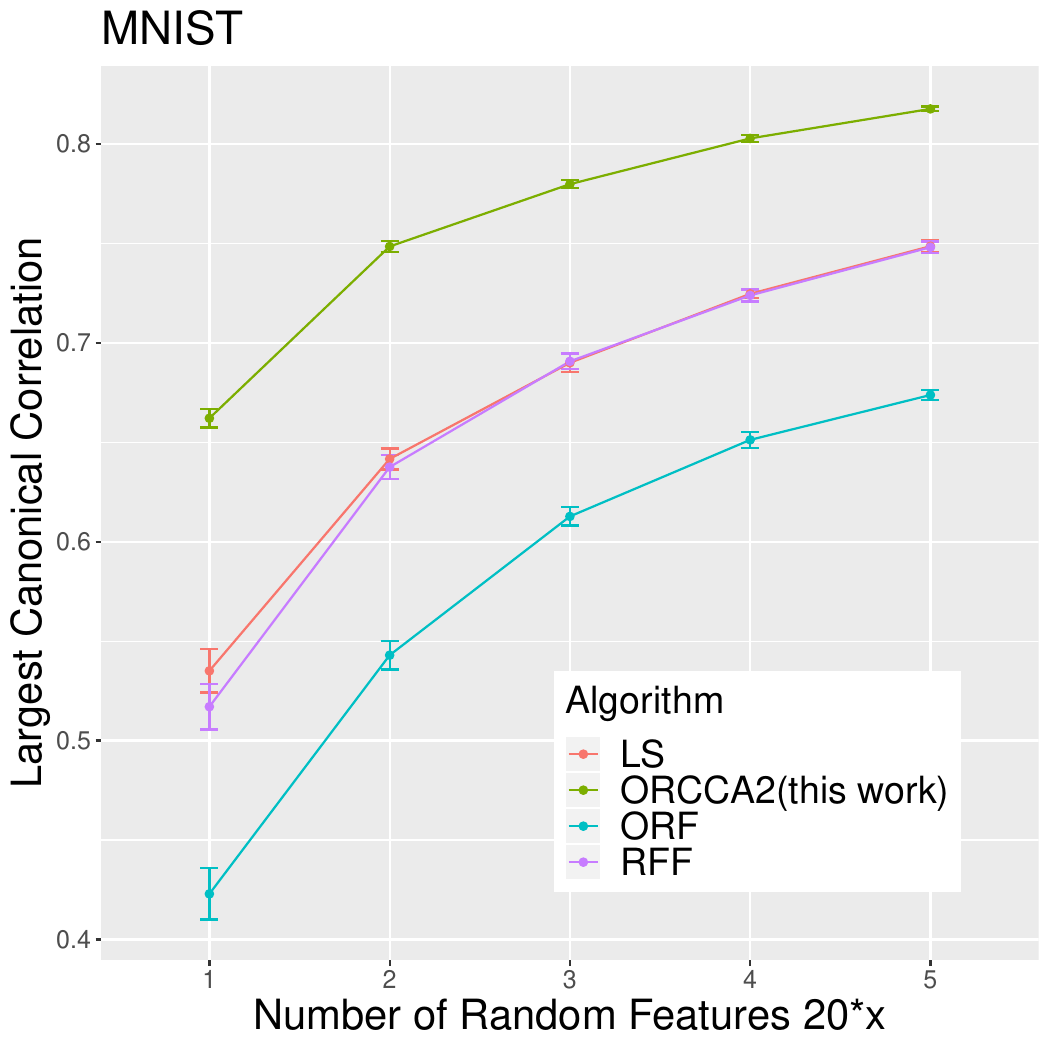}
\caption{The plot of total canonical correlation (first row), top-$10$ canonical correlations (second row), and the largest canonical correlation (third row) versus the number of features obtained by different algorithms. The error bars are obtained with $30$ Monte-Carlo simulations.}
\label{fig: ORCCA2}
\end{figure*}

\begin{figure*}[t!]
\centering
\includegraphics[width = 0.24\textwidth, height=0.13\textheight]{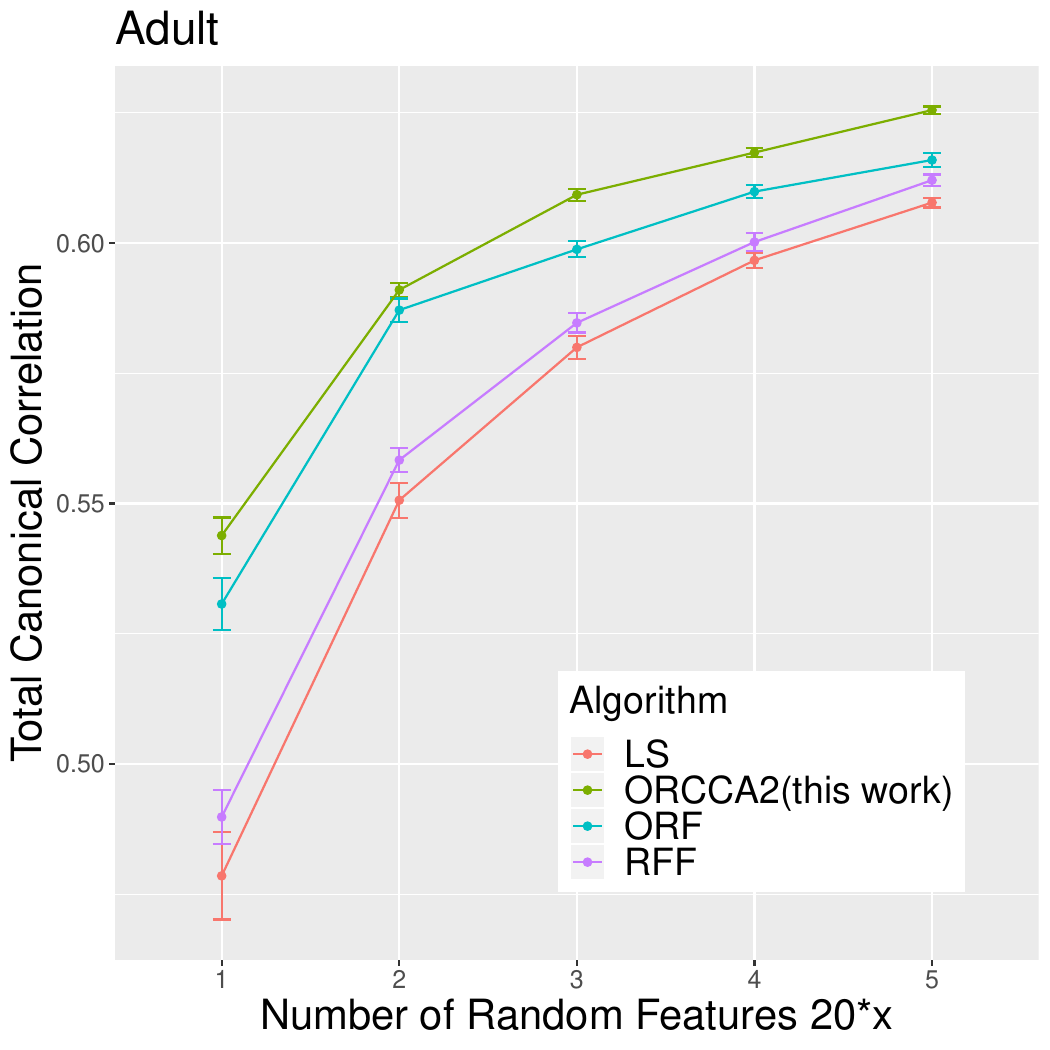}
\includegraphics[width = 0.24\textwidth, height=0.13\textheight]{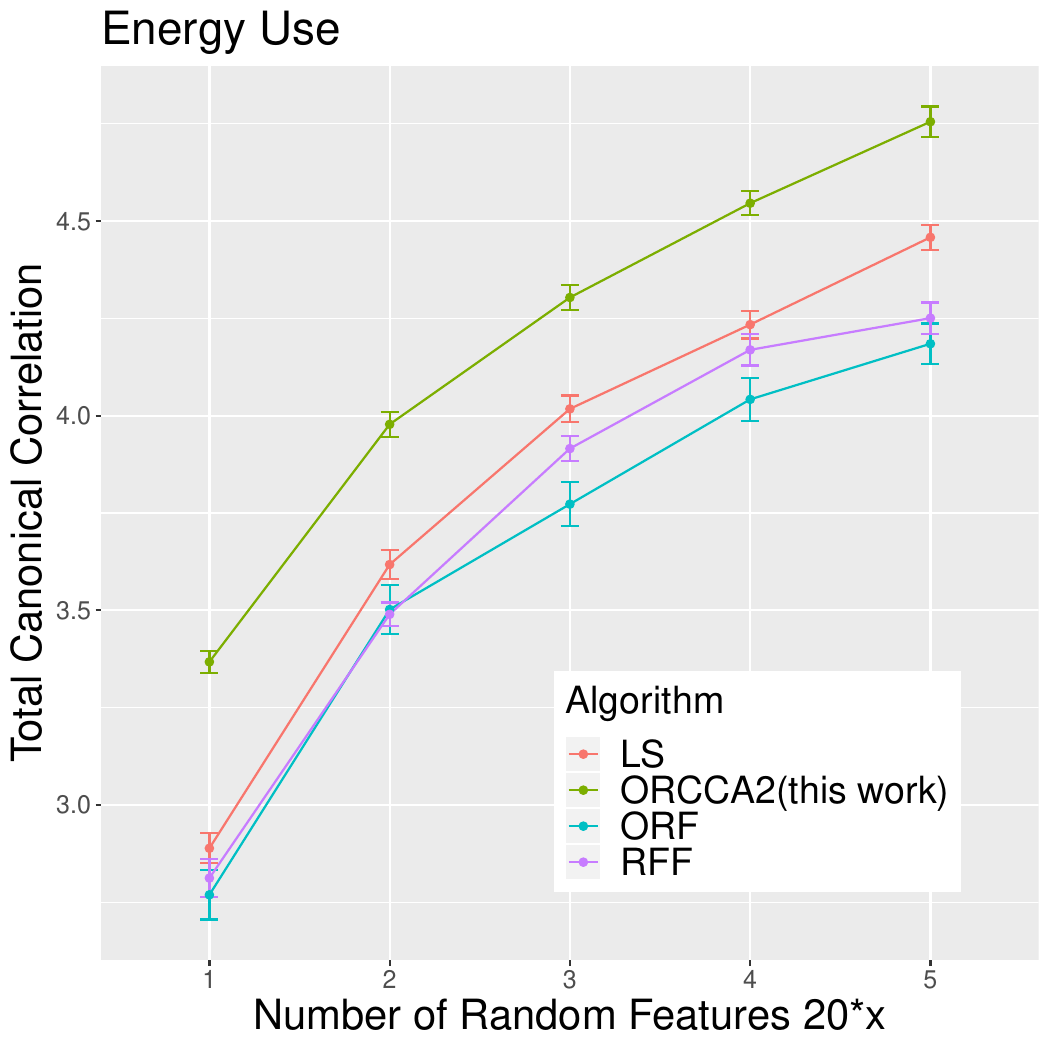}
\includegraphics[width = 0.24\textwidth, height=0.13\textheight]{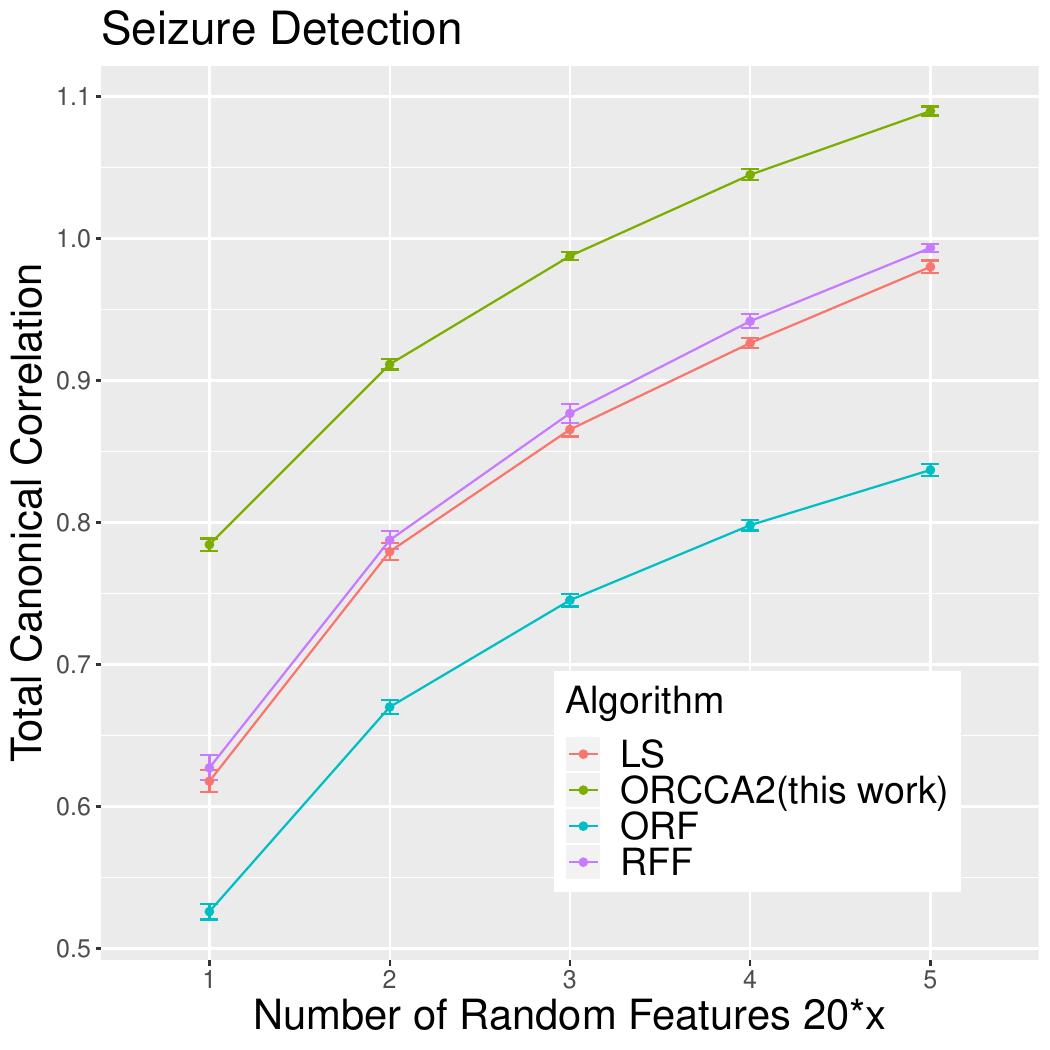}
\includegraphics[width = 0.24\textwidth, height=0.13\textheight]{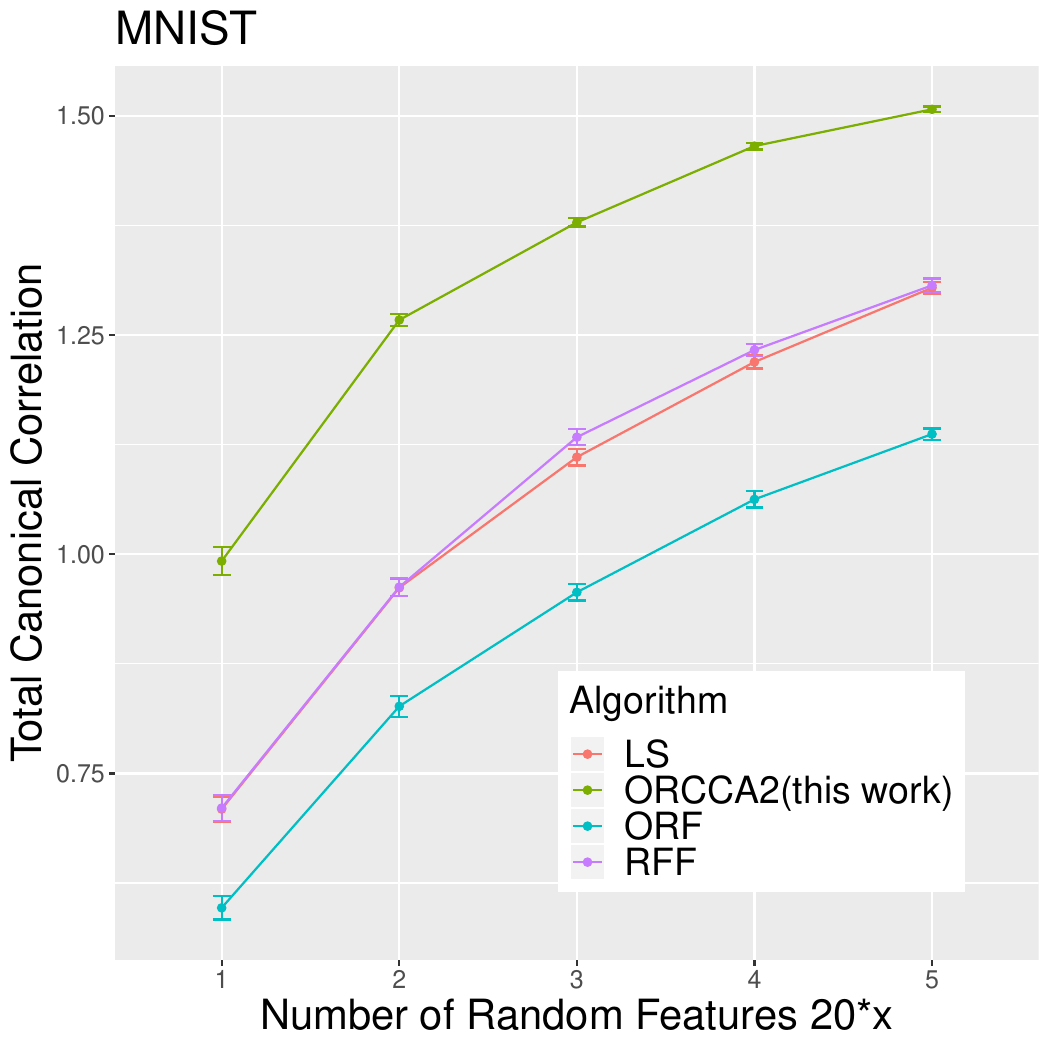}\\
\includegraphics[width = 0.24\textwidth, height=0.13\textheight]{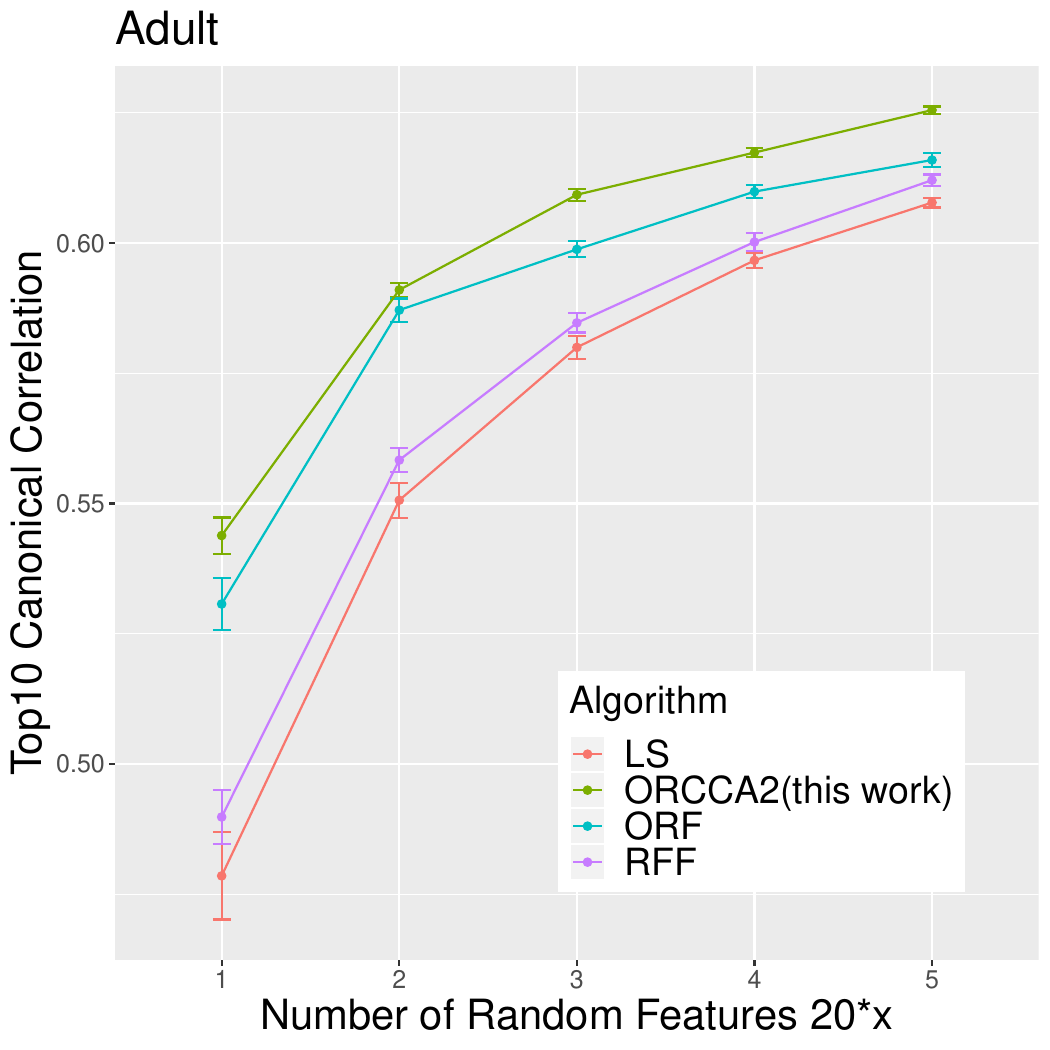}
\includegraphics[width = 0.24\textwidth, height=0.13\textheight]{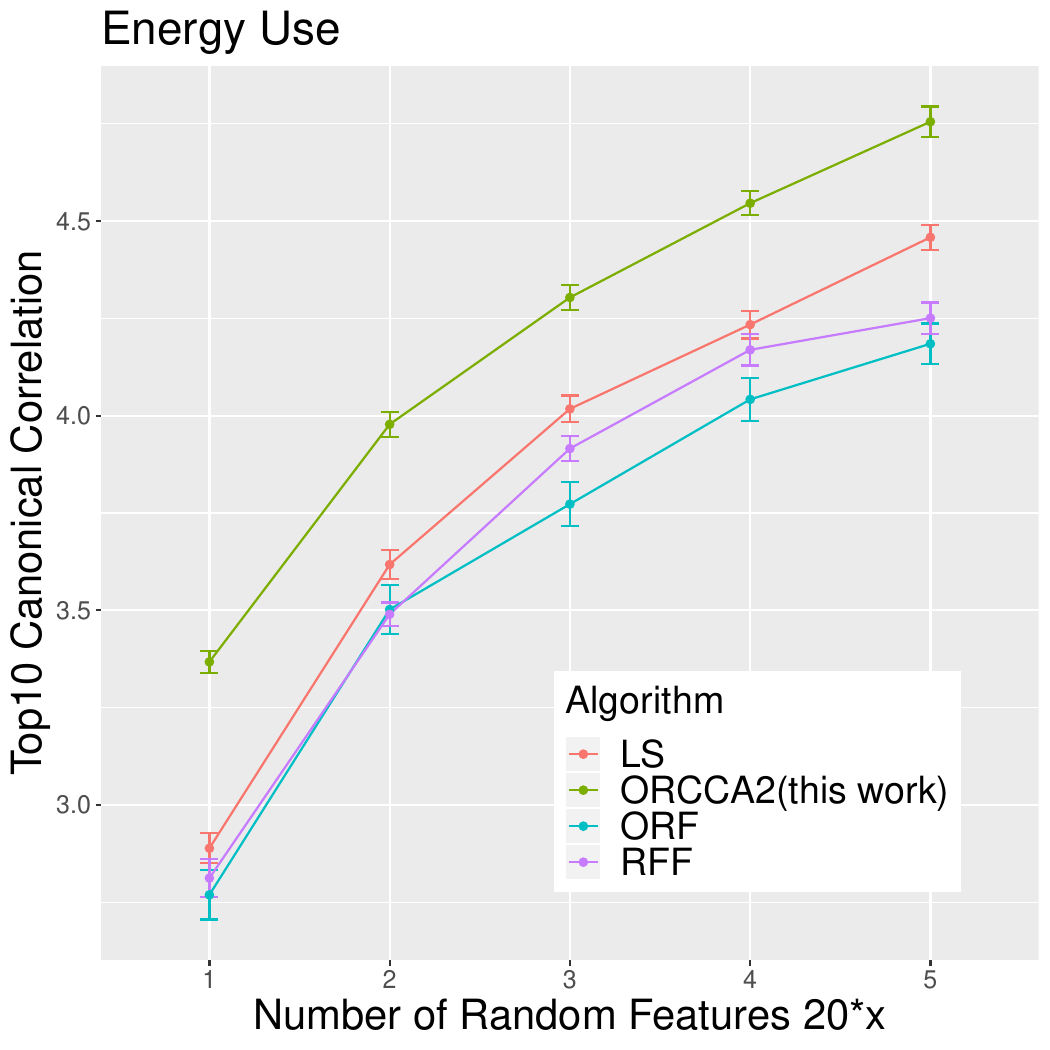}
\includegraphics[width = 0.24\textwidth, height=0.13\textheight]{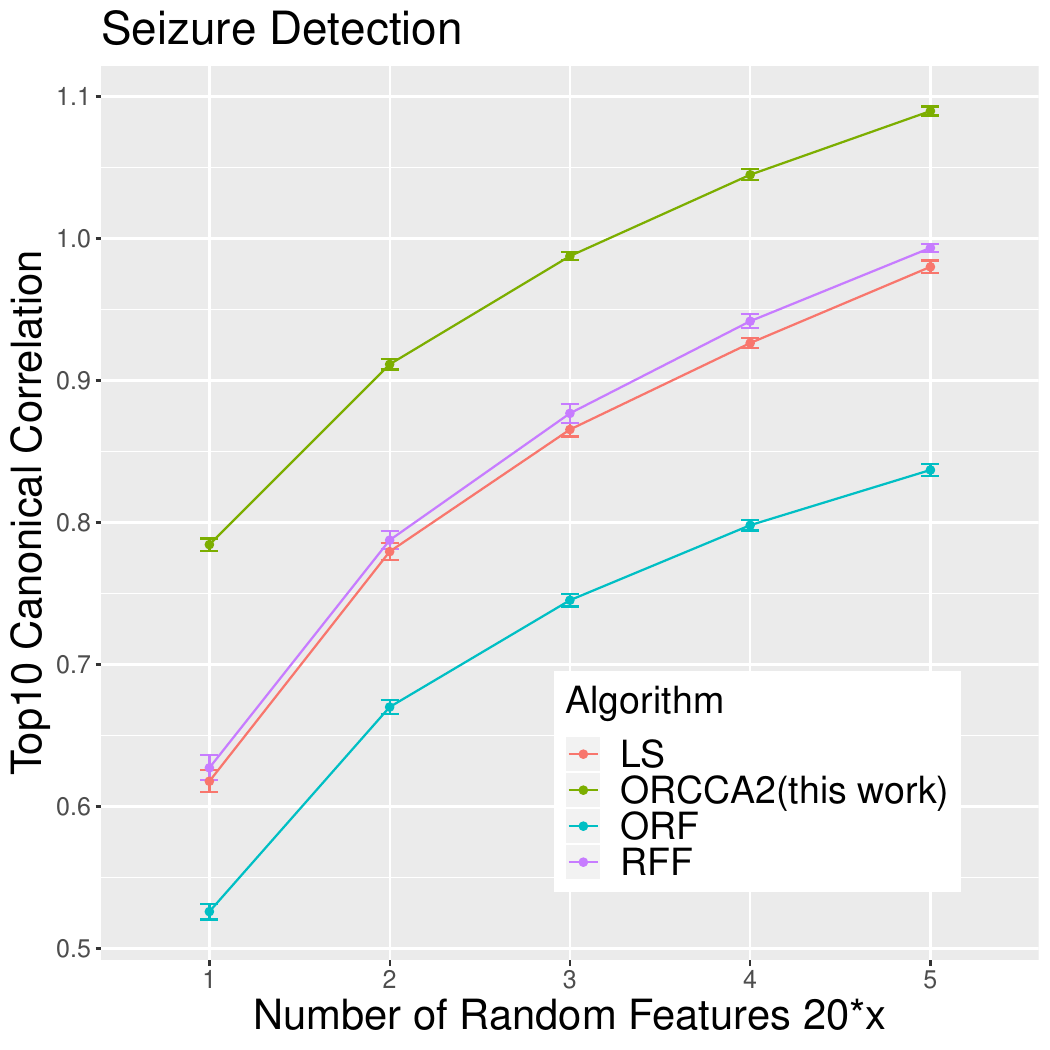}
\includegraphics[width = 0.24\textwidth, height=0.13\textheight]{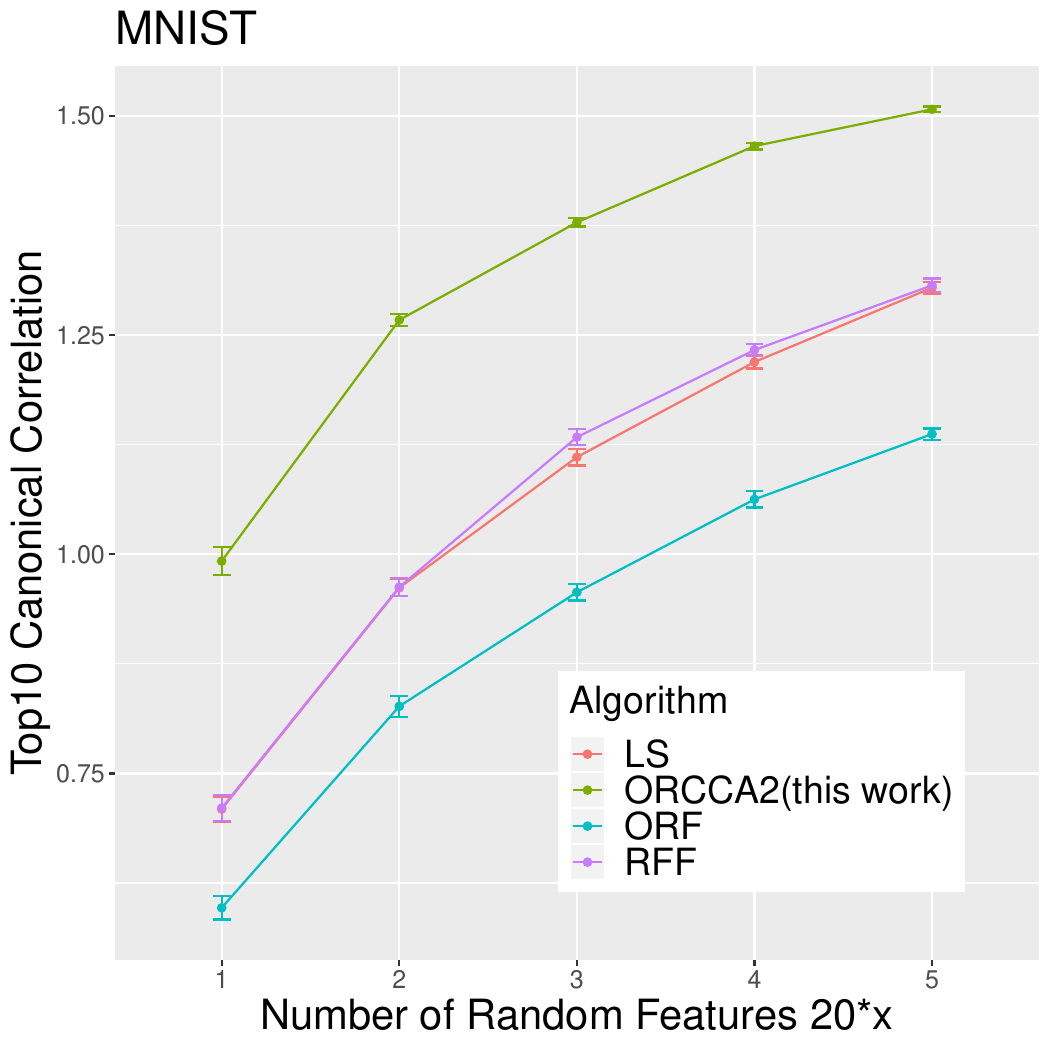}\\
\includegraphics[width = 0.24\textwidth, height=0.13\textheight]{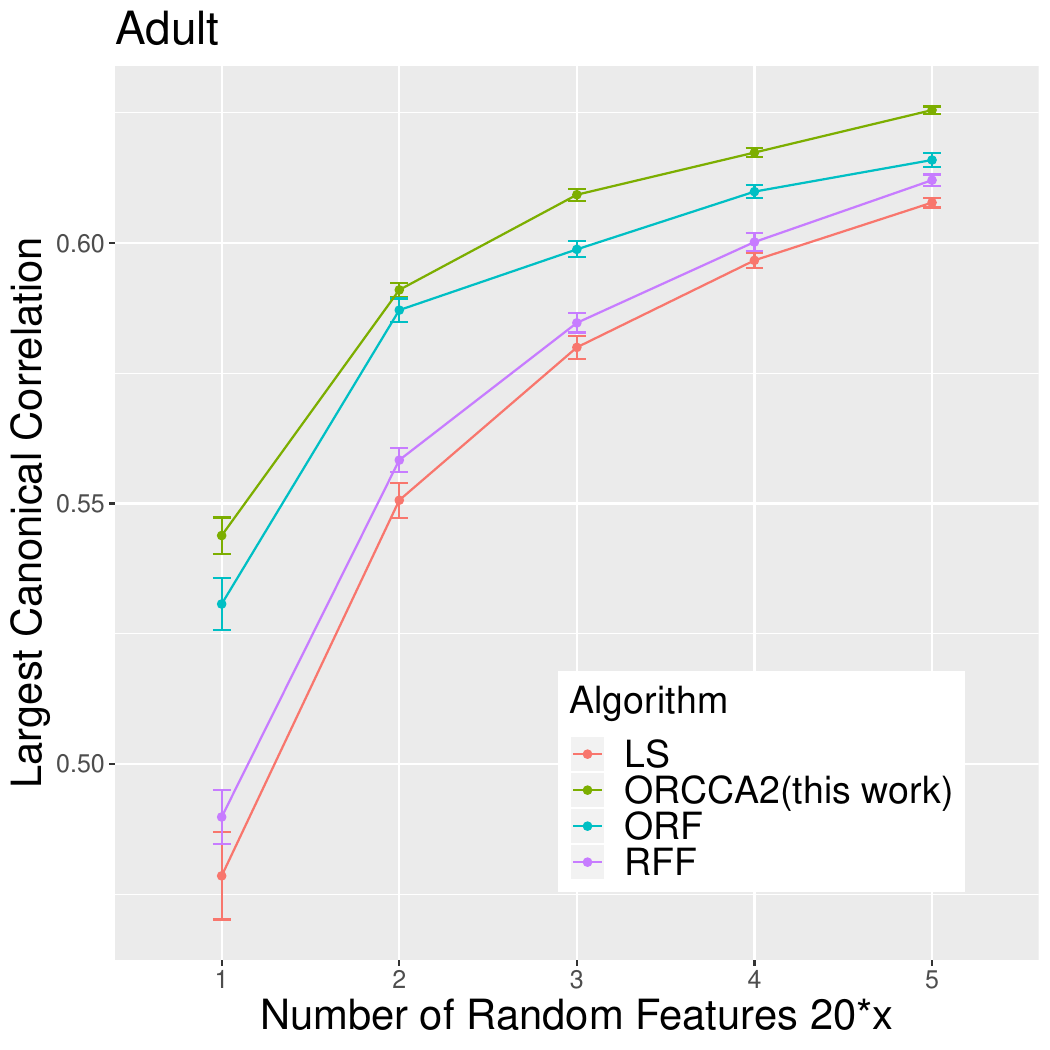}
\includegraphics[width = 0.24\textwidth, height=0.13\textheight]{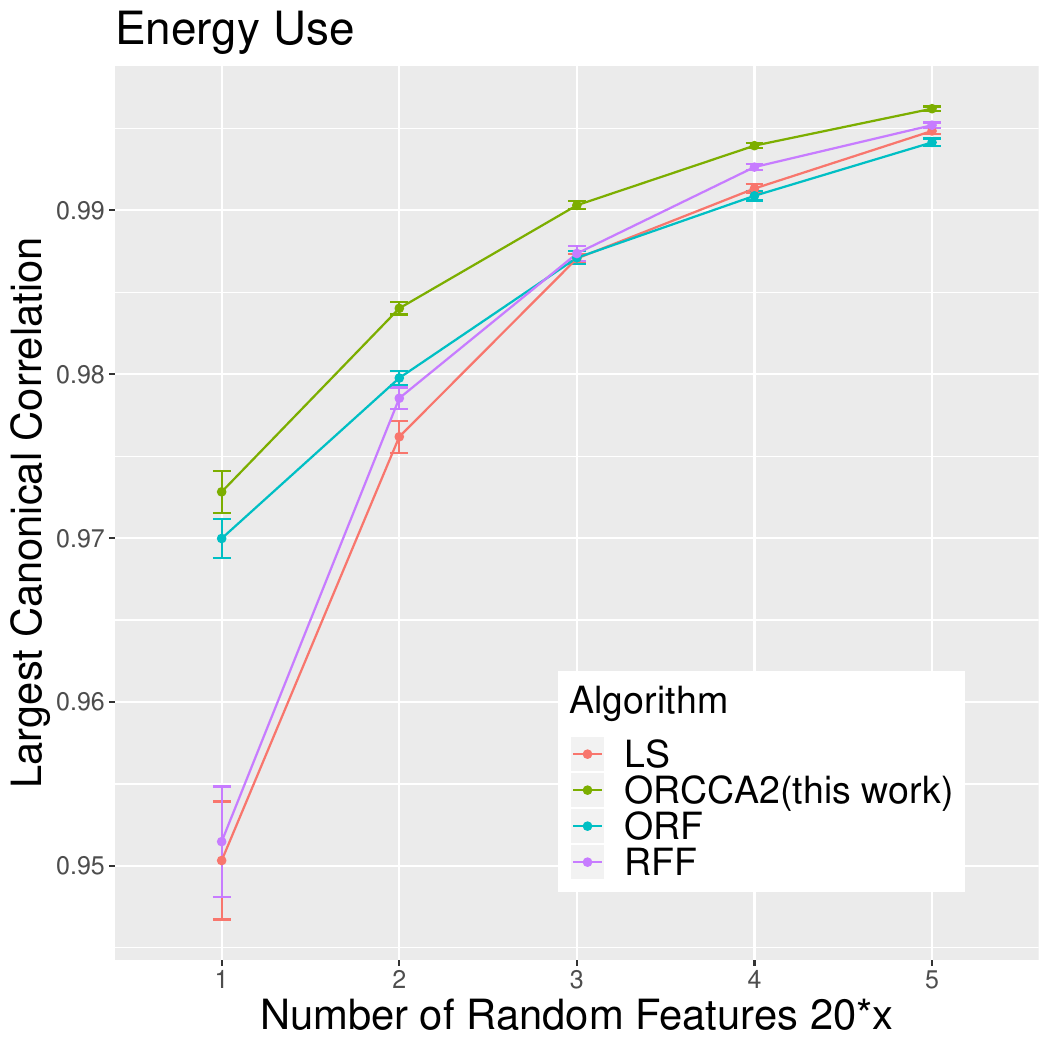}
\includegraphics[width = 0.24\textwidth, height=0.13\textheight]{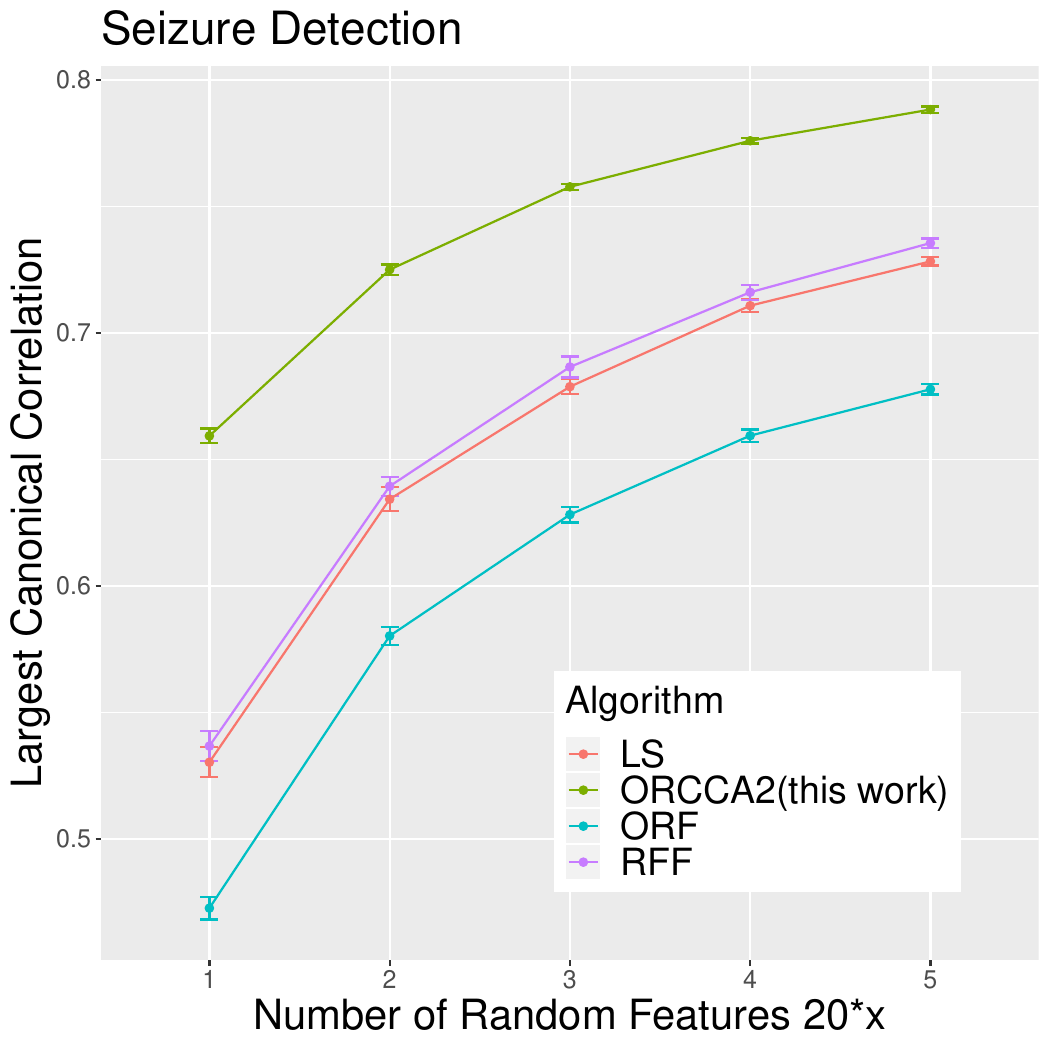}
\includegraphics[width = 0.24\textwidth, height=0.13\textheight]{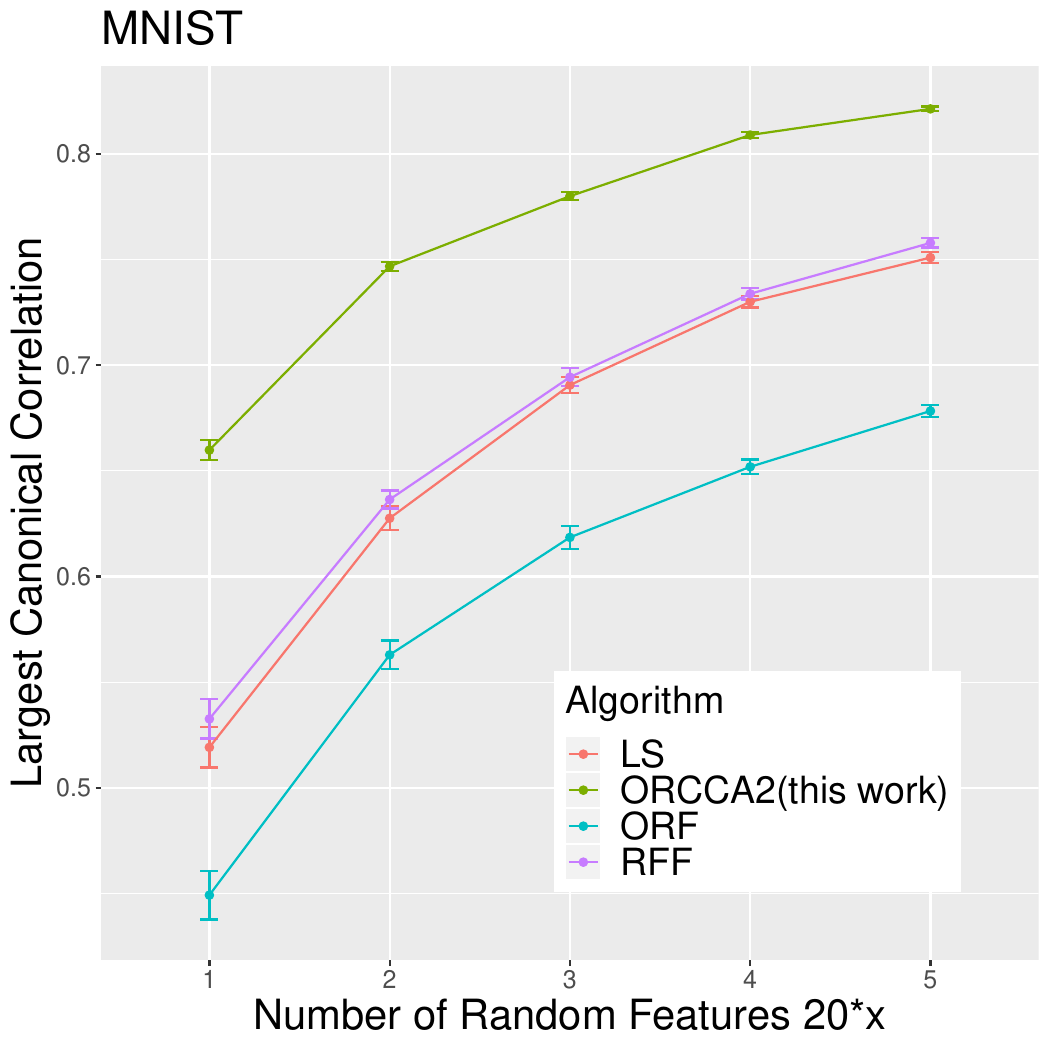}
\caption{The plot of total canonical correlation (first row), top-$10$ canonical correlations (second row), and the largest canonical correlation (third row) versus the number of features obtained by different algorithms on test sets. The error bars are obtained with $30$ Monte-Carlo simulations.}
\label{fig: ORCCA2 with test}
\end{figure*}

\begin{table*}[t!]
\caption{Theoretical time complexity of ORCCA2 and benchmark algorithms and their practical time cost (in sec). For reporting time cost (only), we used $n=5000$ data points, $M=100$ random features, and $M_0=1000$ feature pool. For Energy Use we used the full dataset $n=768$.}
\label{time-table}
\vskip 0.15in
\begin{center}
\begin{small}
\begin{sc}
\begin{tabular}{lcccccr}
\toprule
Algorithms & Complexity            & MNIST  & Energy Use & Seizure Detection & Adult   \\
\midrule
ORCCA2     & $O(n M_0^2 + M_0^3)$  & 21.03  & 2.49       & 17.39             & 15.70   \\
LS         & $O(n M_0^2 + M_0^3)$  & 18.73  & 2.44       & 16.32             & 14.92   \\
RFF        & $O(nM^2)$             & 2.86   & 0.40       & 2.59              & 2.53    \\
ORF        & $O(nM^2)$             & 2.78   & 0.40       & 2.62              & 2.58    \\

\bottomrule
\end{tabular}
\end{sc}
\end{small}
\end{center}
\vskip -0.1in
\end{table*}

\begin{figure*}[t!]
\centering
\includegraphics[width = 0.3\textwidth, height=0.2\textheight]{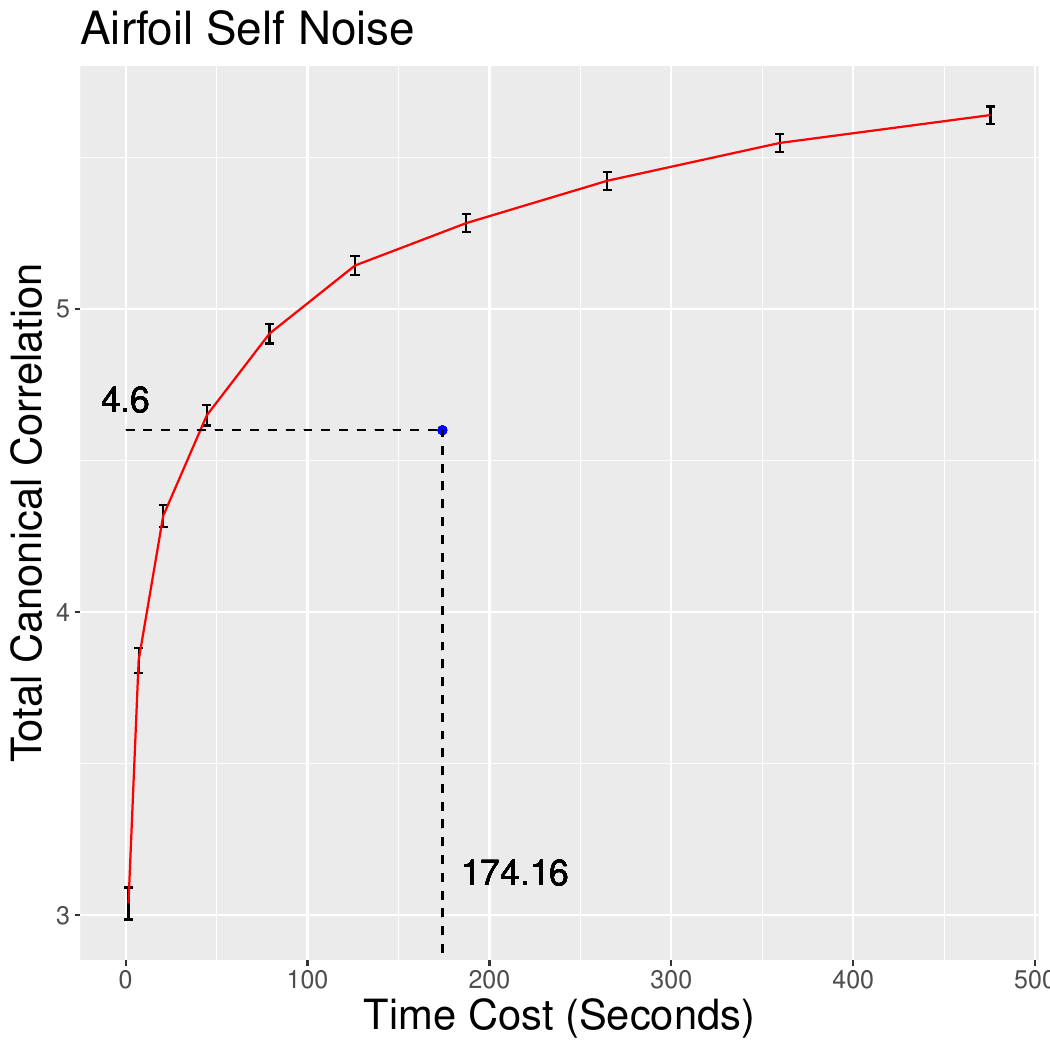}
\includegraphics[width = 0.3\textwidth, height=0.2\textheight]{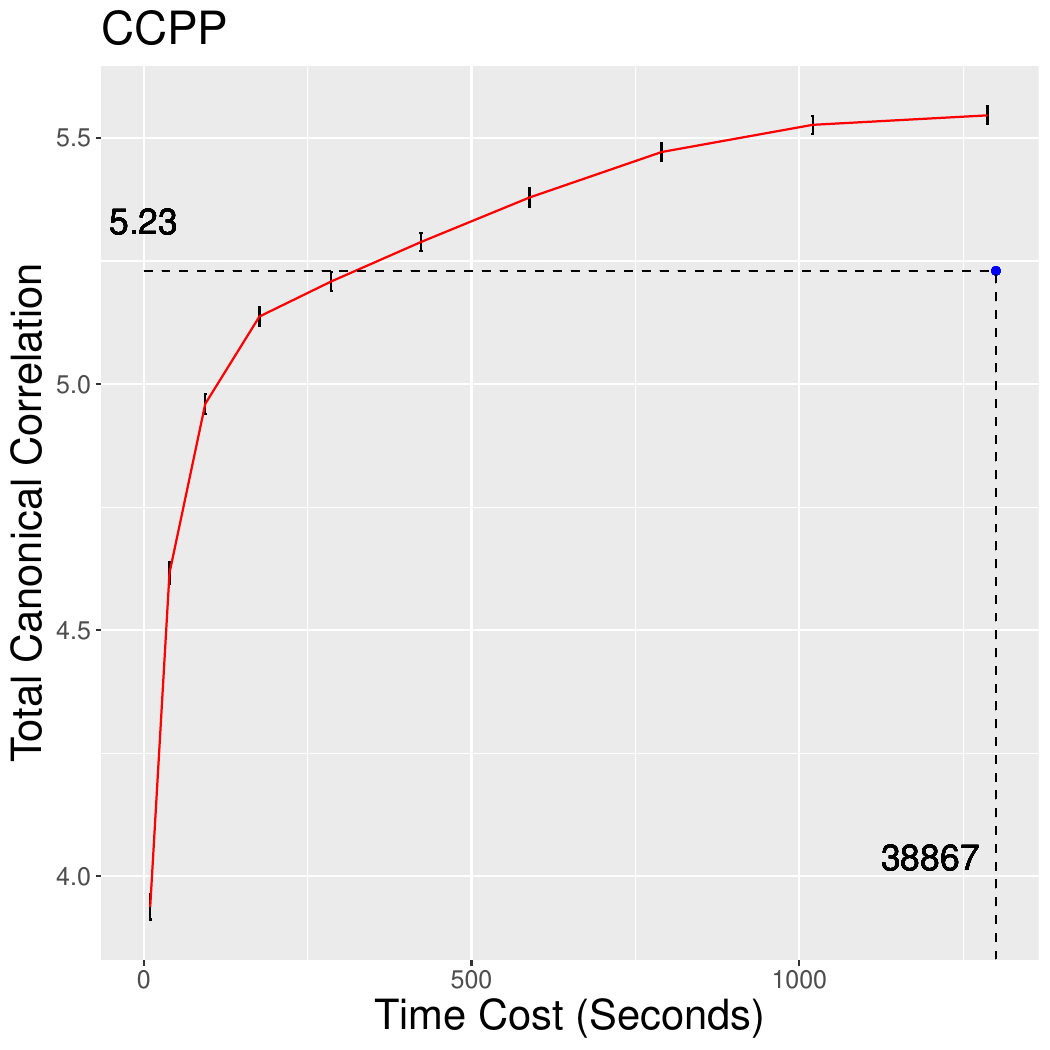}
\includegraphics[width = 0.3\textwidth, height=0.2\textheight]{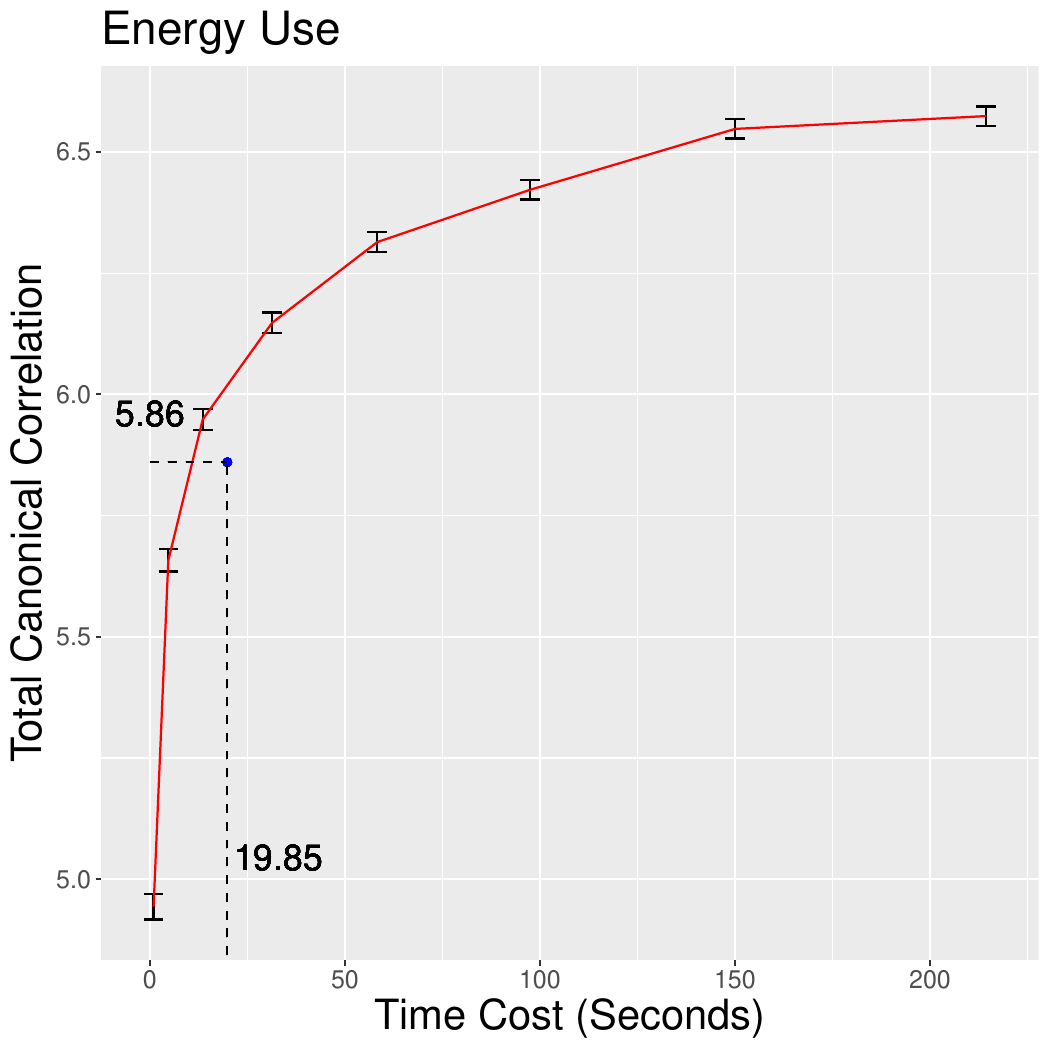}
\caption{ ORCCA2 vs KCCA comparison: The plots represent the total canonical correlations obtained by ORCCA2 with different time costs (by varying the number of random features). The dot represents the total canonical correlation obtained by KCCA and its time cost. The total canonical correlations and time cost of KCCA are also marked by text on the axis. The error bars are obtained with $30$ Monte-Carlo simulations. CCPP is ``Combined Cycle Power Plant'' data.}
\label{kccatable}
\end{figure*}

\section{Numerical Experiments}\label{simulation}

\subsection{Approximated KCCA Comparison}
We now investigate the empirical performance of ORCCA1 and ORCCA2 against other approximated versions of KCCA using six datasets from the UCI Machine Learning Repository.

{\bf Benchmark Algorithms:} We compare our work to four random features based benchmark algorithms that have shown good performance in supervised learning and/or kernel approximation. All four algorithms approximate KCCA by randomized low-rank kernel approximation.
The first one is plain random Fourier features (RFF) \cite{rahimi2008random}. Next is Orthogonal Random Features (ORF) \cite{felix2016orthogonal}, which improves the variance of kernel approximation. We also include two data-dependent sampling methods, LS \cite{avron2017random,bach2017equivalence} and EERF \cite{shahrampour2018data} due to their success in supervised learning as mentioned in Section \ref{sec:relation}. 

{\bf 1) RFF (Random Fourier Features) \cite{rahimi2008random}} with $\phi = \cos(\xb^\top\omegab + b)$ as the feature map to approximate the Gaussian kernel. $\{\omegab_m\}_{m=1}^M$ are sampled from Gaussian distribution $\Nc (0,\sigma^2\Ib)$ and $\{b_m\}_{m=1}^M$ are sampled from uniform distribution $\Uc (0,2\pi)$. We use this to transform $\Xb \in \R^{n\times d_x}$ to ${\Zb}_x \in \R^{n \times M}$. The same procedure applies to $\yb\in \R^n$ to map it to ${\Zb}_y \in \R^{n \times M}$. This algorithm corresponds to the aforementioned RCCA \cite{lopez2014randomized}, and we use RFF instead of its original name to emphasize the role of plain random features method here.

{\bf 2) ORF (Orthogonal Random Features) \cite{felix2016orthogonal}} with $\phi = [\cos(\xb^\top\omegab),\sin(\xb^\top\omegab)]$ as the feature map. $\{\omegab_m\}_{m=1}^{M}$ are sampled from a Gaussian distribution $\Nc (0,\sigma^2\Ib)$ and then modified based on a QR decomposition step. The transformed matrices for ORF are $\Zb_x \in \R^{n \times 2M}$ and $\Zb_y \in \R^{n \times 2M}$. Given that the feature map is 2-dimensional here, to keep the comparison fair, the number of random features used for ORF will be half of other algorithms. 

{\bf 3) LS (Leverage Score Sampling) \cite{avron2017random,bach2017equivalence}} with $\phi = \cos(\xb^\top\omegab + b)$ as the feature map. $\{\omegab_m\}_{m=1}^{M_0}$ are sampled from Gaussian distribution $\Nc (0,\sigma^2\Ib)$ and $\{b_m\}_{m=1}^{M_0}$ are sampled from uniform distribution $\Uc (0,2\pi)$. $M$ features are sampled from the pool of $M_0$ random Fourier features according to the scoring rule of LS \eqref{qelss}. Note that the transformed matrices $\widetilde{\Zb}_x \in \R^{n \times M}$ and $\widetilde{\Zb}_y \in \R^{n \times M}$ correspond to \eqref{trasformed matrix} with $j$-th column normalized by a factor of $\sqrt{p(\omegab_j)/q_{LS}(\omegab_j)}$ due to importance sampling.

{\bf 4) EERF (Energy-based Exploration of Random Features) \cite{shahrampour2018data}} with $\phi = \cos(\xb^\top\omegab + b)$ as the feature map to approximate the Gaussian kernel. $\{\omegab_m\}_{m=1}^{M_0}$ are sampled from Gaussian distribution $\Nc (0,\sigma^2\Ib)$ and $\{b_m\}_{m=1}^{M_0}$ are sampled from uniform distribution $\Uc (0,2\pi)$. $M$ features are selected using the scoring rule in \eqref{qeerf} from the $M_0$ feature pool. We use the sampled $M$ features to transform $\Xb\in \R^{n\times d_x}$ to ${\Zb}_x \in \R^{n \times M}$.

\begin{table*}[t!]
\caption{ORCCA2 performance comparison with other state of the art deep learning variants of CCA. All deep learning algorithms use 20 latent dimensions, RFF and ORCCA2 use 20 random features for a fair comparison. The standard errors are estimated with $30$ Monte-Carlo simulations.}
\label{DLcompare}
\vskip 0.15in
\begin{center}
\begin{small}
\begin{sc}
\resizebox{\textwidth}{!}{
\begin{tabular}{|c|c|c|c|c|c|c|c|c|}
\hline
&\multicolumn{4}{|c|}{FashionMNIST}&\multicolumn{4}{|c|}{MNIST}\\
\hline
Algorithms  & Time(s)  & Largest & Top-$10$ & Total & Time(s) & Largest & Top-$10$ & Total\\
\hline
RFF        & $\mathbf{0.016 \pm 0.001}$ & $0.559 \pm 0.010$ & $3.126 \pm 0.027$ & $3.971 \pm 0.034$ & $\mathbf{0.016 \pm 0.001}$ & $0.405 \pm 0.005$ & $2.773 \pm 0.020$ & $3.586 \pm 0.027$\\
DCCA        & $34.983 \pm 0.020$ & $0.517 \pm 0.026$& $1.993 \pm 0.039$ & $ 2.360 \pm 0.047$ & $33.023 \pm 0.459$ & $0.261 \pm 0.018$&$1.355 \pm 0.056$ & $1.585 \pm 0.072$ \\
DTCCA       &$7.266 \pm 0.422$ & $0.535 \pm 0.008$ & $1.372 \pm 0.028$ & $1.194 \pm 0.042$ & $ 6.911 \pm 0.348$ & $0.173 \pm 0.008$ & $0.731\pm 0.032$ & $0.555 \pm 0.48$\\
DGCCA       &$53.941 \pm 0.171$ & $\mathbf{0.698 \pm 0.014}$ & $1.952 \pm 0.040$ & $2.132 \pm 0.065$ & $55.732 \pm 0.150$ & $0.221 \pm 0.012$ & $1.155 \pm 0.050$ & $1.274 \pm 0.043$\\
DVCCA       &$4.756 \pm 0.018$ & $0.401 \pm 0.006$ & $3.226 \pm 0.055$ & $4.168 \pm 0.122$ & $4.615 \pm 0.010$ & $0.184 \pm 0.004$ & $1.329 \pm 0.034$ & $1.600 \pm 0.042$\\
ORCCA2      &$0.083 \pm 0.003$ & $0.638 \pm 0.008$ & $\mathbf{3.509 \pm 0.021}$ & $\mathbf{4.483 \pm 0.028}$ & $0.090 \pm 0.003$ & $\mathbf{0.452 \pm 0.005}$ & $\mathbf{3.077 \pm 0.017}$ & $\mathbf{4.016 \pm 0.024}$\\
\hline
\end{tabular}}
\end{sc}
\end{small}
\end{center}
\vskip -0.1in
\end{table*}
\begin{enumerate}
\item {\bf Numerical Experiments for ORCCA1}

{\bf Practical Considerations:} Following \cite{lopez2013randomized}, we work with empirical copula transformation of datasets to achieve invariance with respect to marginal distributions. For $\Xb$ domain, the variance of random features $\sigma_x$ is set to be the inverse of mean-distance of $50$-th nearest neighbour (in Euclidean distance), following \cite{felix2016orthogonal}. We use the corresponding Gaussian kernel width for KCCA. The label information $y$ is remained in its original space after coupula transform. For LS, EERF, and ORCCA1, the pool size is $M_0=10M$ when $M$ random features are used in CCA calculation. The regularization parameter $\lambda$ for LS is chosen through grid search. The regularization parameter $\mu=10^{-6}$ is set to be small enough to make its effect on CCA negligible while avoiding numerical errors caused by singularity. The feature map for ORCCA1 is set to be $\phi = \cos(\xb^\top\omegab + b)$. 

{\bf Performance:} The empirical results for ORCCA1 are reported in Fig. \ref{fig: ORCCA1}. The results are averaged over $30$ simulations and error bars are presented in the plots. There is only one canonical correlation to report due to $d_y = 1$. We can clearly observe that ORCCA1 shows dominance over the other benchmark algorithms except for two occasions: (i) at $M=100$ features ORF performs on par with ORCCA1 in Adult dataset and (ii) at $60+$ features EERF performs on par with ORCCA1 in Energy dataset. Another interesting observation here is that other than ORCCA1, all the benchmark algorithms do not show any clear hierarchy in performance given their established empirical performance hierarchy in supervised learning \cite{shahrampour2019sampling}. 

\item{\bf Numerical Experiments for ORCCA2}

{\bf Practical Considerations:} The variance of random features $\sigma_x$ is set to be the inverse of mean-distance of $50$-th nearest neighbour (same procedure as ORCCA1 experiments). After performing a grid search, the variance of random features $\sigma_y$ for $\Yb$ is set to be the same as $\sigma_x$, producing the best results for all algorithms. For LS and ORCCA2, the pool size is $M_0=10M$ when $M$ random features are used in CCA calculation. The regularization parameter $\lambda$ for LS is chosen through grid search. The regularization parameter in CCA calculation remains at $\mu = 10^{-6}$. The feature map for ORCCA2 is also $\phi = \cos(\xb^\top\omegab + b)$.

{\bf Performance:} Our empirical results on four datasets are reported in Fig. \ref{fig: ORCCA2}. The results are averaged over $30$ simulations and error bars are presented in the plots. The first row of Fig. \ref{fig: ORCCA2} represents the total canonical correlation versus the number of random features $M$ for ORCCA2 (this work), RFF, ORF, and LS. We observe that ORCCA2 is superior compared to other benchmarks, and only for the Adult dataset ORF is initially on par with our algorithm. The second row of Fig. \ref{fig: ORCCA2} represents the top-$10$ canonical correlations, where we observe the exact same trend. This result shows that the total canonical correlation are mostly explained by their leading correlations. The third row of Fig. \ref{fig: ORCCA2} represents the largest canonical correlation. Although ORCCA2 is developed for the total canonical correlation objective function, we can still achieve performance boost in identifying the largest canonical correlation, which is also a popular objective in CCA. The theoretical time complexity and practical time cost are tabulated in Table \ref{time-table}. Our cost is comparable to LS as both algorithms calculate a new score function for feature selection. The run time is obtained on a desktop with an 8-core, 3.6 Ghz Ryzen 3700X processor and 32G of RAM (3000Mhz). Given the dominance of ORCCA1 over LS and EERF, and ORCCA2 over LS, we can clearly conclude that data-dependent sampling methods that improve supervised learning are not necessarily best choices for CCA. 
Finally, in Figure \ref{kccatable}, we compare ORCCA2 and KCCA. The datasets used for this part are smaller than the previous one due to prohibitive cost of KCCA. On these datasets, ORCCA2 can gradually outperform KCCA in total CCA value as the computation time increases (due to increasing the number of random features), while it is also more efficient in terms of time cost. The main reason is that ORCCA2 approximates a ``better'' kernel than Gaussian by choosing good features. 

\item{\bf Performance Comparison with Training and Testing Sets (ORCCA2)}

Due to the sample and select nature of ORCCA algorithms, one might wonder the necessity of cross-validation during the implementation of ORCCA. In this experiment, we randomly choose $80\%$ of the data as the training set and use the rest as the testing set. For ORCCA2 and LS, the random features are re-selected according to the score calculated with the training set, and canonical correlations are calculated with these features for the testing set. RFF and ORF algorithms are only implemented on the test set as their choice is independent of data. All the parameters, including the random features variances $\sigma^2_x$ and $\sigma^2_y$, feature map $\phi$, pool size $M_0$, the original distribution for random features, the regularization parameter $\mu$ for CCA, and the regularization parameter $\lambda$ for LS are kept the same as the previous ORCCA2 numerical experiment. The results are shown in Fig. \ref{fig: ORCCA2 with test} following the same layout as Fig. \ref{fig: ORCCA2}. We can observe that the performance of algorithms are almost identical to the results of Fig. \ref{fig: ORCCA2}, which shows the robustness of ORCCA2 to test data as well. In another word, we can avoid the loss of information due to cross-validation when implementing ORCCA algorithms. 

\end{enumerate}

\subsection{Deep Learning CCA Comparison}

\begin{figure}[t!]
\centering
\includegraphics[width = 0.24\textwidth, height=0.2\textheight]{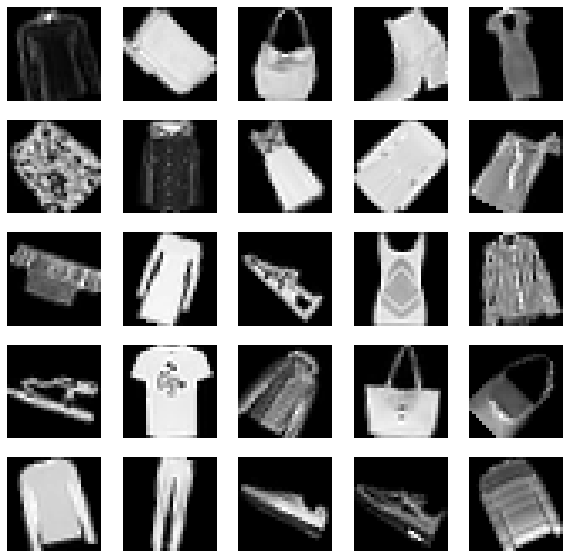}
\includegraphics[width = 0.24\textwidth, height=0.2\textheight]{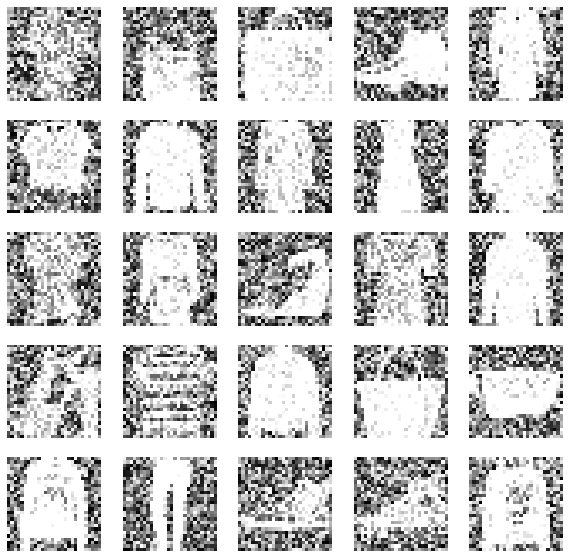}
\includegraphics[width = 0.24\textwidth, height=0.2\textheight]{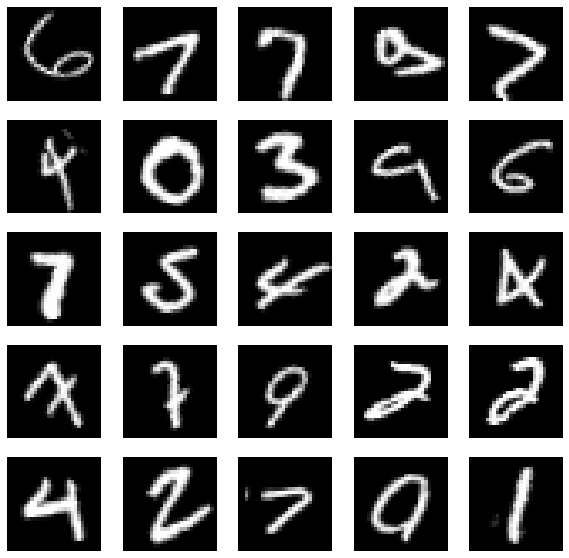}
\includegraphics[width = 0.24\textwidth, height=0.2\textheight]{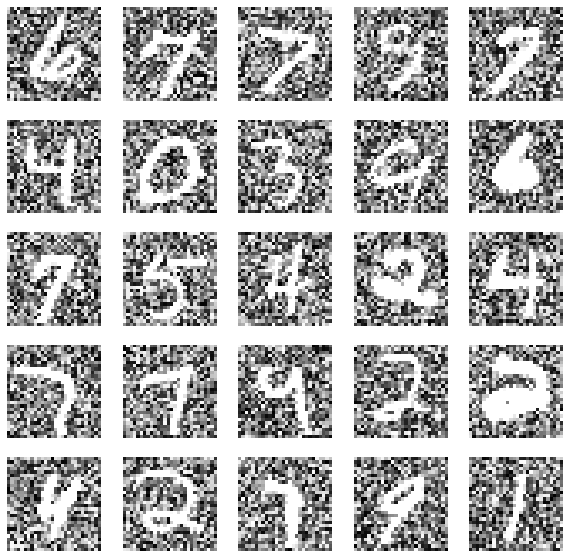}
\caption{Rotated View 1 (left) and Noisy View 2 (right) of FashionMNIST dataset and MNIST dataset.}
\label{fig: datasets}
\end{figure}

In addition to variants of kernel approximation for CCA, recent studies have been utilizing deep learning to further enhance CCA. The recent deep learning algorithms include Deep CCA (DCCA) \cite{andrew2013deep}, Deep Variational CCA (DVCCA) \cite{wang2016deep}, Deep Generalized CCA (DGCCA) \cite{benton2019deep}, and Deep Tensor CCA (DTCCA) \cite{wong2021deep}. In this subsection, we will compare ORCCA2 with the above mentioned state-of-the-art CCA methods.

{\bf Benchmark Algorithms: }
\begin{enumerate}

\item {\bf DCCA:} DCCA uses deep neural networks for the nonlinear transformation of both views of data. The parameters of the network are updated using gradient flow, calculated with negative total canonical correlation used as the loss function. We identically construct the neural networks for both views of the data. The neural network consists of a hidden-layer with $100$ neurons and ReLU activations, with an output of dimension $20$.

\item {\bf DTCCA:} Deep Tensor CCA uses deep neural network for non-linear transformation, similar to DCCA. However, the loss function is replaced by the tensor formulation of total canonical correlations from tensor CCA to extend DCCA to more than two views of data. We construct the same neural network embedding as DCCA. We must note that the main goal of DTCCA is to extend CCA to more than two views of data with tensor formulation. The main reason to include DTCCA in this comparison is to show the impact of multi-view CCA on the correlation extraction.

\item {\bf DGCCA:} Deep Generalized CCA uses deep neural networks for non-linear transformation, similar to DCCA. It forces the two views of data to transform into a shared view that will be learned through gradient flow. We use a similar neural network structure for DGCCA, where we feed the data into a hidden layer with $160$ neurons and ReLU activations. It will then be transformed into a latent variable of dimension $20$. 

\item {\bf DVCCA:} Deep Variational CCA replaces the total canonical correlation loss function with the expected log-likelihood, derived from the evidence lower bound given a latent variable with shared information from both views of data. We use the same network structure as DCCA for DVCCA.

\end{enumerate}
\noindent
{\bf Datasets:} Instead of using low-dimensional datasets, we focus on high-dimensional image data correlation extraction, following the convention in deep learning variants of CCA. We use two datasets, Fashion MNIST, a $28 \times 28$ grayscale image dataset for clothes and shoes, and MNIST, a $28 \times 28$ grayscale image dataset for handwritten digits. For both datasets, we randomly sample $1500$ images, $500$ for training, $500$ for validation, and $500$ for testing. We rotate these images for a degree uniformly distributed in $[-\frac{\pi}{4},\frac{\pi}{4}]$ to form the first view of the data. Then, for each image in View 1, we randomly sample a new image with the same label from the original dataset and add centered Gaussian noise 
to form View 2. This experimental setting is identical to that of \cite{wang2016deep}. The training set in ORCCA2 is used for feature selection, whereas in all deep learning algorithms it is used for neural network training. Validation set is used in all deep learning algorithms for training epochs validation. Test set is then used in all the algorithms for recording canonical correlations.

\noindent
{\bf Practical Consideration:} The latent dimension for deep learning algorithms as well as the number of random features for ORCCA2 and RFF are set to be $20$ for a fair comparison. Note that any deep learning algorithm is able to achieve perfect correlation between two views of training data for large enough latent dimension (over-fitting). However, this is not the intent in applications of nonlinear CCA, and all the deep learning CCA works \cite{andrew2013deep,benton2019deep,wang2016deep,wong2021deep} conduct their experiments in a low latent dimension, as we do here. 
We use ``Adam'' optimizer for all deep learning algorithms, and the number of training epochs is selected using the validation set. DCCA is trained for $50$ epochs. DTCCA is trained for $100$ epochs. DGCCA is trained for $20$ epochs. DVCCA is trained for $300$ epochs. 

\noindent
{\bf Performance:} All the results are tabulated in Table \ref{DLcompare}, where we report the time cost, largest correlation value, the sum of top-$10$ correlation values, and the sum of total correlation values. The standard errors are calculated using $30$ Monte-Carlo simulations. The reported time for deep learning algorithms and ORCCA2 includes the model training/feature selection time in addition to the canonical correlation analysis time. The validation time for deep learning algorithms is not included. As we can observe, ORCCA2 dominates all the benchmark algorithms in correlation extraction, while being significantly faster than the deep learning algorithms.

\section{Conclusion}\label{Conclusion}
Random features have been widely used for various machine learning tasks but often times they are sampled from a pre-set distribution, approximating a fixed kernel. In this work, we highlight the role of the objective function in the learning task at hand. We propose a score function for selecting random features, which depends on a parameter (matrix) chosen based on the specific objective function to improve the performance. We start by drawing connections to score functions for random features in supervised learning. We first show the potential of our score function through a class of dimension reduction and correlation analysis models, which involves a trace operator in their objective functions. We then focus on Canonical Correlation Analysis and derive the optimal score function for maximizing the total CCA. Empirical results verify that random features selected using our score function significantly outperform other state-of-the-art methods for random features. It would be interesting to explore the potential of this score function for other learning tasks as a future direction.

\section*{Appendix}\label{proof}

\subsection{Proof of Proposition \ref{P: PORCCA}}

To find canonical correlations in KCCA \eqref{KCC}, we deal with the following eigenvalue problem (see e.g., Section 4.1. of \cite{hardoon2004canonical})
\begin{equation}\label{CCAsolution}
    (\Kb_x + \mu\Ib)^{-1}\Kb_y(\Kb_y+\mu\Ib)^{-1}\Kb_x\pib_x = \delta^2\pib_x,
\end{equation}
where the eigenvalues $\delta^2$ are the kernel canonical correlations and their corresponding eigenvectors are the kernel canonical pairs of $\Xb$. The solutions to $\yb$ can be obtained by switching the indices of $x$ and $y$. When $\Kb_y$ is a linear kernel, we can rewrite above as follows
\begin{equation*}
    (\Kb_x + \mu\Ib)^{-1}\yb\yb^\top(\yb\yb^\top+\mu\Ib)^{-1}\Kb_x\pib_x = \delta^2\pib_x.
\end{equation*}
Then, maximizing the total canonical correlation (over random features) will be equivalent to maximizing the approximated objective $\tr{(\Kb_x + \mu\Ib)^{-1}\yb\yb^\top(\yb\yb^\top+\mu\Ib)^{-1}\Kb_x}$. Here, we first approximate the $\Kb_x$ at the end of the left-hand-side with $\Zb_x\Zb_x^\top$ such that
\begin{equation}\label{tr1}
    \begin{aligned}
        &\tr{(\Kb_x + \mu\Ib)^{-1}\yb\yb^\top(\yb\yb^\top+\mu\Ib)^{-1}\Kb_x}\\
        \approx &\tr{(\Kb_x + \mu\Ib)^{-1}\yb\yb^\top(\yb\yb^\top+\mu\Ib)^{-1}\Zb_x\Zb_x^\top}\\
        = &\tr{(\Kb_x + \mu\Ib)^{-1}\yb\yb^\top\Zb_x\Zb_x^\top}\frac{1}{\mu + \yb^\top\yb}\\
        \propto &\tr{(\Kb_x + \mu\Ib)^{-1}\yb\yb^\top\Zb_x\Zb_x^\top}\\
        = &\tr{\Zb_x^\top(\Kb_x + \mu\Ib)^{-1}\yb\yb^\top\Zb_x}\\
        =
       &\sum_{i=1}^{M_0}\left[\Zb_x^\top(\Kb_x + \mu\Ib)^{-1}\yb\yb^\top\Zb_x\right]_{ii}\\
       =&\frac{1}{M_0}\sum_{i=1}^{M_0} \zb_x^\top(\omegab_i)(\Kb_x + \mu\Ib)^{-1}\yb\yb^\top\zb_x(\omegab_i),
    \end{aligned}
\end{equation}
where the third line follows by the Woodbury Inversion Lemma (push-through identity). 

Given the closed-form above, we can immediately see that good random features are ones that maximize the objective above. Hence, we can select random features according to the (unnormalized)  $\zb_x^\top(\omegab)(\Kb_x + \mu\Ib)^{-1}\yb\yb^\top\zb_x(\omegab)$. 
Given that random features are sampled from the prior $p(\omegab)$, we can then write the score function as 
\begin{equation}\label{score1}
    q(\omegab) \triangleq p(\omegab)\zb_x^\top(\omegab)(\Kb_x + \mu\Ib)^{-1}\yb\yb^\top\zb_x(\omegab),
\end{equation}
completing the proof of Proposition \ref{P: PORCCA}.

\subsection{Proof of Corollary \ref{C: PORCCA}}
In Corollary \ref{C: PORCCA}, we further approximate \eqref{PORCCA} and to achieve an improved time cost. 

Notice that when we sample $M_0$ random features as a pool, the probability $p(\omegab)$ is already incorporated in the score. Therefore, we just approximate the kernel matrix $\Kb_x$ with $\Zb_x\Zb_x^\top$ (according to \eqref{trasformed matrix}), such that
\begin{equation}\label{qapprox}
    q_x(\omegab_m) \approx  \zb_x^\top(\omegab_m)(\Zb_x\Zb_x^\top+\mu\Ib)^{-1}\yb\yb^\top\zb_x(\omegab_m).
\end{equation}
Then, using the push-through identity again, we derive 
\begin{equation}\label{qh}
    \begin{aligned}
        &\zb_x^\top(\omegab_m)(\Zb_x\Zb_x^\top+\mu\Ib)^{-1}\yb\yb^\top\zb_x(\omegab_m)\\
        =&[\Zb_x^\top(\Zb_x\Zb_x^\top+\mu\Ib)^{-1}\yb\yb^\top\Zb_x]_{mm}\\
        =&[(\Zb_x^\top\Zb_x+\mu\Ib)^{-1}\Zb_x^\top\yb\yb^\top\Zb_x]_{mm}.
    \end{aligned}
\end{equation}
Therefore, for a specific $\omegab_m$, the RHS of \eqref{qapprox} can be rewritten in the following form
\begin{equation*}
    RHS= [(\Zb_x^\top\Zb_x+\mu\Ib)^{-1}\Zb_x^\top\yb\yb^\top\Zb_x]_{mm} = \qh_x(\omegab_m).
\end{equation*}
Then, Corollary \ref{C: PORCCA} is proved.

\subsection{Proof of Theorem \ref{P: DORCCA}}
When $\Yb$ is also mapped into a nonlinear space, we need to work with the eigen-system in \eqref{CCAsolution}. As discussed before, the objective is then maximizing the  approximated version of $\tr{(\Kb_x + \mu\Ib)^{-1}\Kb_y(\Kb_y+\mu\Ib)^{-1}\Kb_x}$. Following the same idea of \eqref{tr1}, we have the following
\begin{equation}\label{decomposition}
    \begin{aligned}
        &\tr{(\Kb_x + \mu\Ib)^{-1}\Kb_y(\Kb_y+\mu\Ib)^{-1}\Kb_x}\\
        \approx & \tr{(\Kb_x + \mu\Ib)^{-1}\Kb_y(\Kb_y+\mu\Ib)^{-1}\Zb_x\Zb_x^\top}\\
        = & \tr{\Zb_x^\top(\Kb_x + \mu\Ib)^{-1}\Kb_y(\Kb_y+\mu\Ib)^{-1}\Zb_x}\\
        =& \frac{1}{M_0}\sum_{m=1}^{M_0} \zb_x^\top(\omegab_m)(\Kb_x + \mu\Ib)^{-1}\Kb_y(\Kb_y+\mu\Ib)^{-1}\zb_x(\omegab_m).\\
    \end{aligned}
\end{equation}
We can then follow the exact same lines in the proof of Proposition \ref{P: PORCCA} to arrive at the following score function
\begin{equation*}
q_x(\omegab)= p_x(\omegab)\zb_x^\top(\omegab)(\Kb_x + \mu\Ib)^{-1}\Kb_y(\Kb_y+\mu\Ib)^{-1}\zb_x(\omegab).
\end{equation*}
The proof for $q_y(\omegab)$ follows in a similar fashion.   
\subsection{Proof of Corollary \ref{C: DORCCA}}
Similar to approximation ideas in the proof of Corollary \ref{C: PORCCA}, we can use the approximations $\Kb_x\approx\Zb_x\Zb_x^\top$ and $\Kb_y\approx\Zb_y\Zb_y^\top$ to get
\begin{equation}
\resizebox{1\hsize}{!}{%
$q_x(\omegab_m) \approx \zb_x^\top(\omegab_m)(\Zb_x\Zb_x^\top + \mu\Ib)^{-1}\Zb_y\Zb_y^\top(\Zb_y\Zb_y^\top+\mu\Ib)^{-1}\zb_x(\omegab_m).$%
}
\end{equation}
Using the push-through identity twice in the following, we have
\begin{equation*}
    \begin{aligned}
        &\zb_x^\top(\omegab_m)(\Zb_x\Zb_x^\top + \mu\Ib)^{-1}\Zb_y\Zb_y^\top(\Zb_y\Zb_y^\top+\mu\Ib)^{-1}\zb_x(\omegab_m)\\
        =&[\Zb_x^\top(\Zb_x\Zb_x^\top + \mu\Ib)^{-1}\Zb_y\Zb_y^\top(\Zb_y\Zb_y^\top+\mu\Ib)^{-1}\Zb_x]_{mm}\\
        =&[(\Zb_x^\top\Zb_x + \mu\Ib)^{-1}\Zb_x^\top\Zb_y(\Zb_y^\top\Zb_y+\mu\Ib)^{-1}\Zb_y^\top\Zb_x]_{mm},
    \end{aligned}
\end{equation*}
which provides the approximate scoring rule, 
\begin{equation}\label{empiricalscore}
\qh_x(\omegab_m) = [(\Zb_x^\top\Zb_x + \mu\Ib)^{-1}\Zb_x^\top\Zb_y(\Zb_y^\top\Zb_y+\mu\Ib)^{-1}\Zb_y^\top\Zb_x]_{mm}.
\end{equation}
The proof for $\qh_y(\omegab_m)$ follows in a similar fashion.

\subsection{Proof of Proposition \ref{upperbound}}
We will demonstrate the proof using the ORCCA2 formulation. Recall the total canonical correlation decomposition derived in \eqref{decomposition}. We will now formally define two quantities below:
\begin{equation}
\resizebox{0.48\textwidth}{!}{$
    \begin{split}
        \rho^{(M)}(\text{RCCA}) &\triangleq \frac{1}{M}\sum_{m=1}^{M}\zb_x^\top(\omegab_m)\Gb\zb_x(\omegab_m)\\
        \rho^{(M,M_0)}(\text{ORCCA})  &\triangleq \underset{\Nc \subseteq [M_0],|\Nc| = M}{\sup}\frac{1}{M}\sum_{m \in \Nc} \zb_x^\top(\omegab_m)\Gb\zb_x(\omegab_m).
    \end{split}
    $}
\end{equation}
where
$$
\Gb\triangleq(\Kb_x + \mu\Ib)^{-1}\Kb_y(\Kb_y+\mu\Ib)^{-1}.
$$

We can see that $\rho^{(M)}(\text{RCCA})$ corresponds to the total canonical correlation obtained by $M$ plain random Fourier features. Similarly, we can see that $\rho^{(M,M_0)}(\text{ORCCA})$ corresponds to the total canonical correlation obtained by ORCCA2, where top $M$ features are selected from a pool of $M_0>M$ plain random Fourier features according to score \eqref{DORCCA}.

Taking expectation over random features, we have that
\begin{equation}
    \begin{split}
        &\E[\rho^{(M,M_0)}(\text{ORCCA})]\\
        =& \frac{1}{M}\E\bigg[\underset{\Nc \subseteq [M_0],|\Nc| = M}{\sup}\sum_{m \in \Nc} \zb_x^\top(\omegab_m)\Gb\zb_x(\omegab_m)\bigg]\\
        \geq& \frac{1}{M} \underset{\Nc \subseteq [M_0],|\Nc| = M}{\sup} \E\bigg[\sum_{m \in \Nc} \zb_x^\top(\omegab_m)\Gb\zb_x(\omegab_m)\bigg]\\
        \geq& \frac{1}{M}\E\bigg[\sum_{m \in [M]} \zb_x^\top(\omegab_m)\Gb\zb_x(\omegab_m)\bigg]\\
        =& \E[\rho^{(M)}(\text{RCCA})]=\rho(\text{KCCA}).
    \end{split}
\end{equation}
The above equation shows that the total canonical correlation obtained by ORCCA2 provides an upperbound for the total canonical correlation obtained by RFF in the view of $\Xb$. We can conclude the same thing in the view of $\Yb$ following the same procedure. Setting $\Kb_y = \yb\yb^\top$, we can also have the same property for ORCCA1. Therefore, Corollary \ref{upperbound} is proved.

\subsection{Computation Complexity}\label{comp}

To highlight the computational advantage, we examine the computational cost of the dominating components in KCCA formulation and ORCCA formulation. 
\begin{itemize}
    \item For KCCA, the dominating component in terms of computational cost is the matrix inversion of the full rank matrices $\Kb_x + \mu\Ib$ and $\Kb_y + \mu\Ib$, as we can see in \eqref{KCC}. Both of them have the time complexity of $\Oc(n^3)$ as $\Kb_x,\Kb_y \in \R^{n\times n}$. The matrix multiplication will also induce a theoretical time complexity of $\Oc(n^3)$. Therefore, the computational cost of KCCA is $\Oc(n^3)$.
    \item For ORCCA1 and ORCCA2, the dominating component in terms of computational cost comes from calculating the empirical score function \eqref{empiricalscore}. The score function needs to be computed once for ORCCA1 and twice for ORCCA2. The matrix inversion of $\Zb_x^\top\Zb_x + \mu\Ib$ induces a cost of $\Oc(M_0^3)$, where $M_0$ is the number of random features in the pool. The matrix multiplication $\Zb_x^\top\Zb_x$ induces a cost of $\Oc(nM_o^2)$. Therefore, the computation cost of ORCCA1 and ORCCA2 is dominated by this term and the cost is $\Oc(nM_0^2 + M_0^3)$.
\end{itemize}

\section*{Acknowledgments}
The authors gratefully acknowledge the support of NSF Award \#2038625 as part of the NSF/DHS/DOT/NIH/USDA-NIFA Cyber-Physical Systems Program. The authors submitted the first version of the manuscript at Texas A\&M University. They gratefully acknowledge the support of Texas A\&M University as well as Texas A\&M Triads for Transformation (T3) Program.

\bibliographystyle{./IEEEtran}
\bibliography {references.bib}

\end{document}